\documentclass{article}

\usepackage{arxiv}

\usepackage[utf8]{inputenc} % allow utf-8 input
\usepackage[T1]{fontenc}    % use 8-bit T1 fonts
\usepackage{hyperref}       % hyperlinks
\usepackage{url}            % simple URL typesetting
\usepackage{booktabs}       % professional-quality tables
\usepackage{amsfonts}       % blackboard math symbols
\usepackage{nicefrac}       % compact symbols for 1/2, etc.
\usepackage{microtype}      % microtypography
\usepackage{lipsum}
\usepackage{authblk}

\usepackage{cite}
\usepackage{natbib}
\setcitestyle{authoryear}
\usepackage{balance}
\usepackage{soul}
\usepackage{pifont}
\usepackage{amsthm}
\usepackage{amsmath, amsfonts}%, amssymb}
\usepackage{mathtools}
\usepackage{multirow}
\usepackage{subcaption}
\usepackage{algorithm}
\usepackage{algorithmicx}
\usepackage[noend]{algpseudocode}
\usepackage{comment}
\usepackage{enumitem}
\usepackage{xspace}
\usepackage{epigraph}
\newcommand*{\eg}{\textit{e.g.,}\@\xspace}
\newcommand*{\ie}{\textit{i.e.,}\@\xspace}
\newcommand*{\vs}{\textit{vs.}\@\xspace}
\newcommand*{\wrt}{\textit{w.r.t.}\@\xspace}
\makeatletter
\newcommand*{\etc}{%
	\@ifnextchar{.}%
	{\textit{etc}}%
	{\textit{etc.}\@\xspace}%
}
\makeatother

\algtext*{EndWhile}
\algtext*{EndFor}
\algtext*{EndIf}
\algdef{SE}[DOWHILE]{Do}{doWhile}{\algorithmicdo}[1]{\algorithmicwhile\ #1}%

\newtheorem{proposition}{Proposition}

\makeatletter
\def\BState{\State\hskip-\ALG@thistlm}
\makeatother

\usepackage{xcolor}

\newcommand\numberthis{\addtocounter{equation}{1}\tag{\theequation}}

\usepackage[textsize=scriptsize,backgroundcolor=green]{todonotes}
\title{{\em Same Same, But Different}: Conditional Multi-Task Learning for Demographic-Specific Toxicity Detection}

\author[1,4]{\textbf{Soumyajit Gupta}$^*$}
\author[1,4]{\textbf{Sooyong Lee}$^*$}
\author[2,4]{\textbf{Maria De-Arteaga}}
\author[3,4]{\textbf{Matthew Lease}}
\affil[1]{Department of Computer Science}
\affil[2]{McCombs School of Business}
\affil[3]{School of Information}
\affil[4]{The University of Texas at Austin}
\affil[ ]{\texttt{\{smjtgupta,sooyonglee\}@utexas.edu}, \texttt{dearteaga@mccombs.utexas.edu}, \texttt{ml@utexas.edu}}

\begin{document}

\maketitle

\def\thefootnote{*}\footnotetext{These authors contributed equally to this work}
\def\thefootnote{\arabic{footnote}}

\begin{abstract}

Algorithmic bias often arises as a result of {\em differential subgroup validity}, in which predictive relationships vary across groups. For example, in toxic language detection, comments targeting different demographic groups can vary markedly across groups. In such settings, trained models can be dominated by the relationships that best fit the majority group, leading to disparate performance. We propose framing toxicity detection as multi-task learning (MTL), allowing a model to specialize on the relationships that are relevant to each demographic group while also leveraging shared properties across groups. With toxicity detection, each task corresponds to identifying toxicity against a particular demographic group. However, traditional MTL requires labels for all tasks to be present for every data point. To address this, we propose \textbf{Cond}itional \textbf{MTL} (CondMTL), wherein only training examples relevant to the given demographic group are considered by the loss function. This lets us learn group specific representations in each branch which are not cross contaminated by irrelevant labels. Results on synthetic and real data show that using CondMTL improves predictive recall over various baselines in general and for the minority demographic group in particular, while having similar overall accuracy. 

\end{abstract}

\section{Introduction}
\label{sec:intro}

%\sg{general TL intro}With the massive volume of data generated in social media platforms, the prevalence of hate speech and offensive language, which we jointly refer to as toxic language, in such posts have increased as well. Since manual review of all generated data is infeasible, there has been growing interest in the use of 
In developing natural language processing (NLP) models to detect toxic language \citep{schmidt2017survey,arango2019hate,vaidya2020empirical}, we typically assume that toxic language manifests in similar forms across different targeted groups. For example, HateCheck \citep{rottger2021hatecheck} enumerates templatic patterns such as ``I hate [GROUP]'' that we expect detection models to handle robustly across groups. Moreover, we typically pool data across different demographic targets in model training in order to learn general patterns of linguistic toxicity across diverse demographic targets. 

However, the nature and form of toxic language used to target different demographic groups can vary quite markedly.
% \ml{do we want an example? table 7?}
% \sg{maria said no, so lets skip}
Furthermore,  an imbalanced distribution of different demographic groups in toxic language datasets risks over-fitting %can lead to an implicit, predominant modeling of 
forms of toxic language most relevant to the majority group(s), potentially at the expense of systematically weaker model performance on minority group(s). For this reason, a ``one-size-fits-all'' modeling approach may yield sub-optimal performance and more specifically 
%systematically underperform for one or more groups targeted by toxic language, 
raise concerns of algorithmic fairness \citep{park-etal-2018-reducing,arango2019hate,sap2019risk}. 
%\ma{here we can cite something--ideally work on bias in hate speech that considers this specific issue}. 
At the same time, radically siloing off datasets for each different demographic target group would prevent models from learning broader linguistic patterns of toxicity across different demographic groups targeted. To characterize this phenomenon in which toxic language exhibits both important commonalities and important differences, we borrow the popular phrase of ``\textit{same same, but different}'' \citep{wotton_2019}. 

More formally, such heterogeneity of toxic language targeting different groups can be conceptually framed in terms of  \textit{differential subgroup validity} \citep{hunter1979differential}: a relationship $f: X \rightarrow Y$ mapping the input data $X$ to labels $Y$ may have different predictive power across groups. %, without considering potential disagreement on the labels $Y$.  
The wide diversity of demographics targeted by toxic language, and the ways in which minority groups may be disproportionately targeted, underscores the importance of understanding and recognizing this phenomenon. % in this context. 

From an algorithmic fairness perspective, it has been shown that excluding sensitive attributes from the features used in prediction, also known as
%On one hand, such modeling violates 
 ``fairness through unawareness'' is ineffective~\citep{Dwork.2012}. Some methods, \eg adversarial fairness approaches~\citep{zhang2018mitigating}, address this problem by penalizing models from learning relationships that are predictive of sensitive attributes. 
%{\em anti-classification}, 
%advocates against modeling protected attributes like race or gender %from being explicitly used by a model to make predictions 
%in order to promote fairness, 
%(sometimes referred to as ). 
%However, 
Others have noted that making use of such attributes may significantly improve performance for minority groups and reduce algorithmic bias~\citep{lipton2018does,Kleinberg.2018b}, a reason that is tightly linked to the presence of differential subgroup validity. Prior work \citep{corbett2018measure} % in the fairness literature 
has shown that differential subgroup validity can be addressed by training models that learn group-specific idiosyncratic patterns, such as decoupled classifiers~\citep{dwork2018decoupled}. %One of the reasons why this happens is because if
%Without demographic attributes, trained models can be dominated by the relationships that best fit the majority group, which can lead to disparate performance over other sub-groups. 
In the context of toxic language detection, inclusion of demographics has the potential to boost performance in detecting toxic language targeting the minority group(s) who are less represented in a given dataset.   

%this highlights the importance of explicitly considering demographic attributes to ensure that posts targeting minority groups are correctly labeled by the model.

To address the challenge of differential subgroup validity in toxicity detection, we propose to model %the SSBD nature of 
demographic-targeted toxic language via multi-task learning (MTL). %\footnote{The TL (toxic language) acronym is unrelated to the ``TL'' in MTL (multi-task learning).} \citep{ruder2017overview,crawshaw2020multi}.
%\ma{the confusion between MTL and TL is another reason to drop TL entirely}
MTL combines shared and task-specific layers, allowing a model to specialize on relationships relevant to different groups while leveraging shared properties across groups. In this setting, each MTL {\em task} %in the model 
corresponds to detecting toxic language targeting a different group. %Consider the task of TL detection for the same post, but given two demographic groups (A and B): a) given the post targets group A, is it toxic? and b) given the post targets group B, is it toxic?
%\ma{I commented out the sentence on "same post, different groups because it suggests flawed counterfactuals and is at odds with what we've said about differential subgroup validity}
Shared layers can 
%In a MTL setup, we want the shared structure to 
benefit from training across posts targeting multiple groups, while task-specific layers are trained only on posts that target each respective group. %(or set of groups)\sg{not sure if set of group makes sense here. Matt, did you mean groups which are combined due to preprocessing?}. 
For \eg if a post targets group $A$, it should influence the shared layers and its own task-specific layers, but not task-specific layers for group $B$. 

A key limitation of prior MTL work \citep{liu2019fuzzy} is the assumption that each training point is labeled for all tasks. In our context, this means that a post targeting group $A$ is labeled as non-toxic for group $B$. However, this risks contaminating group $B$ data because toxic terms or phrases contained in such posts would then be associated with a non-toxic label. We refer to this as \textit{label contamination}.

% This can be problematic because it 

% \textbf{Trad}itional \textbf{MTL} (TradMTL)

% Many MTL works exist around this domain
% \ma{which domain? toxicity detection?} which are mostly extensions of architectures or loss functions from the computer vision community \citep{ruder2017overview} (see \textbf{Section \ref{sec:litrev}}). We categorize all these works as \textbf{Trad}itional \textbf{MTL} (TradMTL). However, TradMTL assumes that each training point has labels for all tasks. This means that a post targeting group A would simply be assumed to be non-toxic for group B. This can be problematic, because it  %Conceptually, since Task B is conditioned upon posts targeting that group, it violates the basic framing of the task to train on posts that do not target the group.
% \ma{I tweaked this because I think that is a valid (albeit less effective) conceptualization, so I am framing ours as a different conceptualization in the following paragraph} risks contaminating group B data, since any toxic terms or phrases contained in the post would be associated with a non-toxic label. We refer to this issue as \textit{label contamination} (see \textbf{Section \ref{sec:contam}}).

In this work, we propose a novel  \textbf{Cond}itional \textbf{MTL} (CondMTL) approach that is capable of learning group specific representation of toxic language over sparse labels. Crucially, we abandon the traditional MTL assumption that each training instance is labeled for all tasks. Instead, instances labeled only for one task influence shared-layer training without contaminating training of task-specific layers for other tasks. 
%each group-specific task only considers posts targeting that group. %, and the label indicates whether it is toxic or not. 
%For example, if groups correspond to gender, then an example of a non-toxic instance in the women-specific branch would be "women are great", whereas posts targeting other genders would not receive a label for this branch\ma{I added an example for clarity, since "target" has a negative connotation, people may not understand what we are doing immediately}. In order to achieve this, we propose an updated labeling schema that allows the NLP model to identify 
We achieve this by introducing a task-specific, conditional representation of labels, combined with a novel conditional loss function for MTL training that is a selective variant of the widely-used binary cross-entropy loss. % measure for classification. 

%examples that are relevant to its own branch and not be contaminated by labels from other groups, and propose a different loss function to enable learning from sparse labels\ma{check that you like the characterization as "propose a different loss function", I borrowed wording from contributions}. 

% \ma{We should emphasize two things: (1) domain-specific: heterogeneity across groups, importance of performing well for minorities (maybe cite lit emphasizing harms to minorities)} 

% This differential perception\ma{I wouldn't center perceptions} of the same content by different sub-groups is referred in literature as differential subgroup validity \ma{this is not what diff. subgroup validity refers to. It refers to the predictive relationship (f: X->Y) being different across groups, without considering potential disagreement on Y} [CITE]. 

To evaluate CondMTL for toxic language detection with differential subgroup validity, 
%To empirically study the problem of differential subgroups validity in this context, and assess both the risks of label contamination and the benefits of the proposed CondMTL approach, 
we use the {\em Civil Comments} \citep{borkan2019nuanced} portion of the {\em Wilds} \citep{koh2021wilds} dataset, which provides toxicity labels with target demographics.
%This dataset has demographic targeted annotations for gender groups \etc and toxicity labels for the comments\ma{I think a question we can expect is why did we only test it for gender (why?). I would skip this and just say we use this dataset and the gender labels--> connect the use of gender labels to differential subgroup validity and having a majority and minority subgroup}. In this work we consider two gender groups of men and women for illustration of results. In the case of toxic language detection, since the prevalence of toxicity is usually low (see \textbf{Section \ref{sec:data}}), it is necessary that the NLP classifier has better predictive performance in detecting toxic examples \ie higher recall for this label. Our experiments (see \textbf{Section \ref{sec:res}}) demonstrate that using our proposed CondMTL along with the updated group labeling schema results in 
%Results show improved performance of toxic language detection for both demographic groups\ma{"both demographic groups"-> you have not said which are "both", since you haven't said you focus on gender} when compared with other MTL model variants. 
%Our r
%In evaluating toxic language detection, it is useful to 
We center the potential harms affecting the targets and distinguish {\em misses} (\ie an undetected toxic post) {\vs} {\em false alarms} (\ie a non-toxic post that is erroneously flagged) because the potential harm of different types of errors is different. Recall, which is tightly linked to the algorithmic fairness measure of Equality of Opportunity~\citep{hardt2016equality}, is thus a central measure of interest in our study. Our results show that CondMTL matches accuracy of baselines while boosting recall across demographic target groups. For CondMTL, we also observe improved Equality of Opportunity between the groups. Further analysis investigates how the group specific nature of our problem affects model predictions (Section \ref{sec:disc}). We also present analysis of model performance \vs a prior baseline (Section \ref{sec:ana}) and conduct additional synthetic experiments to verify model correctness (Appendix \ref{app:benchmark}). Finally, we show that CondMTL reduces time and space required \vs other MTL baselines.

\vspace{0.5em}
\noindent\textbf{Key contributions} of our work are as follows:
\begin{itemize}[leftmargin=*] % looks weird
%\begin{itemize}
\item We propose a novel Conditional MTL framework that supports conditional labels in general and specifically addresses differential subgroup validity in the context of toxic language detection. % \wrt target demographics.
\item Results show that CondMTL performs better than existing MTL variants across demographic groups considered.
% over the toxic labels across all groups, even in an imbalanced dataset.
%\item To our knowledge, we are the first to apply this conditional logic in the domain of TL detection\ma{have others have done it in other domains? if so, can we claim we are "proposing a novel loss function"?}. While traditional MTL permits both the contamination of weights in the shared layer and erroneous backpropagation of losses in the task-specific branches, CondMTL amends these problems.
\item We provide theoretical and empirical justification with benchmarking scenarios for mathematically checking model validity.
%\item Our CondMTL can substitute standard loss measure without any change in existing architecture, facilitating easy adoption of this logic in future works. 
\item For reproducibility and adoption, we share our TensorFlow-based source code for CondMTL\footnote{\bf \url{https://github.com/smjtgupta/CondMTL}}.%\hl{We share our sourcecode} for reproducibility\st{, with a one-line code change to facilitate ease of adoption.} 
\end{itemize}

\section{Related Work} \label{sec:litrev}

Multi-task learning (MTL) has widespread use in many applications of machine learning, from computer vision to natural language processing and from hate speech detection to media bias diagnosis. We present a focused overview of the existing MTL works as applicable to our study. For detailed analysis, readers are referred to  surveys by \citep{ruder2017overview}, \citep{crawshaw2020multi}, and \citep{vandenhende2021multi}. 

MTL architectures broadly fall under two categories of a) hard; and b) soft parameter sharing of hidden layers. Hard parameter sharing shares the hidden encoder layers between all tasks while keeping several task-specific output layers. It is the most commonly used approach \citep{kokkinos2017ubernet,long2017learning,kendall2018multi} to MTL and empirically reduces the risk of over-fitting \citep{baxter1997bayesian}. Soft parameter sharing maintains several independent task-specific layers with either some form of regularization in loss \citep{yang2017deep} or introduces correlation units across tasks \citep{misra2016cross}. It can learn an optimal combination of shared and task-specific representations as shown in \citep{gao2019nddr,liu2019end,ruder2019latent}.

% \newpage

\vspace{-0.5em}
\subsection{MTL in Computer vision} 

% Methods often share convolutional and fully-connected layers; these approaches vary with a combination of task specific layers, dimension reduction unit or regularization over loss function. 

\textbf{Works in Architecture.} \citep{long2017learning} place matrix priors on the fully connected layers, allowing the MTL model to learn relationships between tasks. They rely on a pre-defined structure for sharing, which may be well-studied for computer vision problems but less adaptable for non-vision tasks. \citep{lu2017fully} follows a bottom-up approach starting with thin layers and dynamically widens them greedily during training by promoting grouping of similar tasks. This greedy grouping may inhibit the model from learning similarities between tasks. Cross Stitch Networks \citep{misra2016cross} maintain correlation units across tasks specific layers, called cross stitch units. These units use a linear combination of the output of the previous layers across tasks, allowing the units to learn appropriate weights if the tasks are correlated to each other. 

\textbf{Works in Loss Function.} These methods do not alter the MTL architecture, but rather use modified loss functions. \citep{kendall2018multi} use a Gaussian likelihood with a task-dependant uncertainty inspired loss function that assigns relative weights to each task based on its difficulty. \citep{chen2018gradnorm} constrains the task-specific gradients to be of similar magnitude, thereby forcing all tasks to learn at an equal trajectory. \citep{guo2018dynamic} assigns higher weights to difficult tasks, forcing the network to learn better values on these tasks. However, this requires manual parameter tuning. \citep{liu2019end} aims to maintain equal trajectory in losses across tasks, irrespective of their difficulty, by readjusting weights across tasks after each gradient update. Tasks with slower drop rate are assigned more weights for the next update.

\vspace{-0.5em}
\subsection{MTL in Natual Language Processing}

In order to find better hierarchies across NLP tasks, methods are developed to account for losses, not only at the outermost layer of the model, but at intermediate layers as well. \citep{sogaard2016deep} empirically show that using supervision losses at the earlier stages of an NLP pipeline improved model performance for tasks such as part-of-speech tagging and named entity recognition on a deep bi-directional RNN architecture. \citep{hashimoto2017joint} uses the aforementioned idea to apply losses at different levels of the hierarchical NLP pipeline. They propose a joint model for MTL with supervising losses at the word, syntactic, and semantic levels.

% \begin{remark}
% Note that all the strategies described before are heuristic based, where the improvement expectation is only empirical in nature and not guaranteed theoretically or otherwise over different datasets and tasks. Our proposed CondMTL is theoretically justified and empirically shown to achieve better performance in the prescribed problem setting. See Results (Section \ref{sec:res}) and Analysis (Section \ref{sec:ana}).
% \end{remark}

\subsection{MTL in Hate Speech}

\citep{liu2019fuzzy} use a fuzzy MTL setup for hate speech detection to identify hate speech from single-labeled data. They employ a fuzzy rule based schema to identify potential groups of hate targets over examples and update the rule thresholds \wrt training error. For hate speech, typically the hate class is the smaller class with fewer examples, so they mark all unlabelled examples in their dataset as the larger class \ie non-hate, which leads to the issue of \textit{label contamination} (see Section \ref{sec:contam}). %\ma{The sentence here did not connect to the rest, so I commented it out. if you want to bring it back, make sure you explain how it connects because right now it comes across like a random detail of their data processing} 
\citep{plaza2021multi} observe that invoking labels of sentiment, emotion, and target of hate speech jointly improves detection of hate speech. They employ a MTL model where each tweet has labels corresponding to the mentioned attributes, using transfer learning. \citep{samghabadi2020aggression} develop a MTL model to jointly predict aggression and misogyny across datasets. \citep{vaidya2020empirical} use a MTL model to jointly predict toxicity and identity of target. \citep{morgan-etal-2021-wlv} frame the identification of toxic, engaging, and fact-claiming comments on social media as a MTL problem. \citep{awal2021angrybert} jointly learns hate speech detection with sentiment
classification as primary and target identification as secondary task, and employ a model with independent task-specific layers and a parallel shared layer. The output vectors from these layers are merged using a gate fusion mechanism, which is a linear combination unit between the shared and task specific vectors. %Their MTL model performs empirically better for some tasks over multiple datasets.

% \vspace{-1.5em}

\vspace{-0.5em}
\subsection{Applied MTL} 

Given the media's ability to shape individuals' actions and beliefs, prior work has sought to improve media bias detection to identify underlying biases and their influences. \citep{spinde2022exploiting} use a MTL model which learns to detect bias using task-specific layers associated with specific datasets, yielding performance that %. \citeauthor{spinde2022exploiting} observe that this model 
generally surpasses their baseline single-task methods. The proliferation of misinformation has driven prior work in automated fact-checking. \citep{vasileva2019takes} use a MTL model to learn from multiple fact-checking organizations simultaneously. Their MTL model yields sizeable improvements over their single-task learning baseline, indicating a benefit to jointly learning identification of fact-check worthy claims for multiple news sources. %Members within a demographic group can behave differently from one another and efforts to categorize demographics can lead to undesirable generalizations. 
In the context of data annotation, collected labels may often be dominated by views of the majority. To address this, \citep{gordon2022jury} introduce jury learning to model individual labelers in the dataset and incorporate dissenting voices when forming juries of diverse demographics. 

Variations in language across posts targeting different groups is at the core of our motivation. We aim to improve model performance and fairness by learning group-specific variations. As we explained in Section~\ref{sec:intro}, this contrasts with approaches such as adversarial fairness, which aim to \emph{prevent} a model from relying on group-specific variations. For instance, 
%Closest to our work is research by 
\citep{huang2019neural} argue that there is no unified use of language across different demographics including gender, age, country and region. With the goal of improving generalization across groups, they propose an adversarial training approach; % to guide the error in the MTL model to account for demographic variations. 
the adversarial branch predicts demographics of the document, while the main branch does standard text classification. In doing this, their objective is to learn patterns that generalize across groups. Meanwhile, our goal is to learn group-specific patterns, \eg the model should be able to recognize demeaning terms that are only used against one minority group. %They empirically show that this joint training helps reduce classification errors across demographics, across four datasets. 

\vspace{-0.25em}
\section{Problem Statement}

We are given a dataset $\mathcal{D} \in \mathbb{R}^{N \times F}$, with $N$ samples (posts) and $F$ features. These $F$ dimensional features can be extracted using any off-the-shelf NLP model. We assume binary labels for this dataset as $\mathcal{Y} \in \mathbb{R}^N$, where each label ($y$) corresponding to a post ($d$) can be either Non-Toxic (0) or Toxic (1). Furthermore, we are given the $K$ demographic groups pertaining to the targets of each post. Thus each post can be mapped to an overall (group-agnostic) toxicity label as well as multiple group-targeted toxicity labels as $d \rightarrow y$ and  $d \rightarrow y_k, \forall k \in K$,%\ma{I am confused by the notation of s and k, isn't k the group? what is s?}
where the overall label $y=\bigcup_K y_k$ considering the data to be toxic/non-toxic, irrespective of the group. Due to the nature of the $K$ independent groups, we have the combined dataset $\mathcal{D}={\mathcal{D}_1 \cup \mathcal{D}_2 \cup \ldots \cup \mathcal{D}_{K}}$ as the union of the demographic specific data points $\mathcal{D}_k \in \mathbb{R}^{N_k \times F}$. 

If our objective is to optimize a certain performance metric, %maximize accuracy\ma{but is this our objective? we may want to stay away from framing accuracy as the objective in the problem statement, since this contradicts what we say is important in the domain} 
\ie minimizing Binary Cross Entropy (BCE), we can do it over: a) the entire dataset $\mathcal{D}$; or b) demographic group specific data subsets $\mathcal{D}_k, k \subseteq K$. The Single Task (STL) model has independent classifiers for each split $\mathcal{D}_k$, while the Multi Task Learning (MTL) variant has one joint classifier with $K$ task-specific branches for each split $\mathcal{D}_k$.

\section{Conditional MTL} \label{sec:condmtl}

Conditional loss is intuitive: for each demographic branch, we should compute error across toxic and non-toxic class labels only for examples that are relevant to that branch's demographic group. 

\subsection{Labeling Schema} \label{sec:contam}

Traditional MTL (TradMTL) approaches assume that each training point has labels for all tasks. A post targeting group {\color{teal} Green}, irrespective of toxicity label, is assumed to be non-toxic towards group {\color{orange} Orange} as well. This formulation of the task leads to many posts containing toxic language being labeled as non-toxic, by the labels marked red as shown in \textbf{Table~\ref{tab:contam_tweet}}. % violate the basic framing of the task
%\ma{(I edited to address this point)I think this is a different framing of the task, rather than a violation (one can frame the task as "is this toxic against group G", which is a valid and intuitive task, albeit a less effective one). I would position our contribution as a new framing, rather than claiming the labeling schema others have proposed is fundamentally wrong. This both (1) emphasizes our contribution at the conceptual level, (2) avoids painting other work in an unfairly negative light}
We argue that this labeling schema, which blends together the questions \emph{is the post toxic?} and \emph{who is the target of the post?}, leads to \textit{label contamination}. For extended illustration, refer to Appendix \ref{app:contam}.

\begin{table}[ht]
    \vspace{-0.5em}
    \centering
    \resizebox{0.7\linewidth}{!}{%
    \begin{tabular}{c|cc|cc} \toprule
        Post & \multicolumn{2}{c|}{Traditional MTL Labels} & \multicolumn{2}{c}{Correct Labels} \\ 
        & {\color{teal} Green} & {\color{orange} Orange} & {\color{teal} Green} & {\color{orange} Orange} \\ \midrule
        ``I hate {\color{teal} Green}'' & Toxic & {\color{red} Non-Toxic} & Toxic & {\color{blue} $\bullet$}\\
        ``I love {\color{teal} Green}'' & Non-Toxic & {\color{red} Non-Toxic} & Non-Toxic & {\color{blue} $\bullet$} \\ \bottomrule
    \end{tabular}
    }
    \caption{\small Label contamination occurs in a Traditional MTL label assignment when
    posts that target a given group ({\color{teal} Green}) are assumed to be non-toxic toward any other group (\eg {\color{orange} Orange}). 
    {\color{red}Red} denotes unsupported label assignments, while ({\color{blue} $\bullet$}) correctly denotes  
    that these posts do not contain a label wrt.\ the {\color{orange} Orange} target group.}
    \vspace{-1.5em}
    \label{tab:contam_tweet}
\end{table}

%The conditional logic has a stark distinction from the traditional labeling schema in \citep{liu2019fuzzy}, where demographically non-relevant examples were categorized as non-toxic as well. 
%\ma{I edited this paragraph a fair bit, to see track changes see version commented out in its original placement}
In order to let the model differentiate between demographic-specific examples, we consider group-conditional labels from the set $\{$T, NT, $\bullet\}$, where $\bullet$ is an indicator denoting that the label of the current example is irrelevant/unknown \wrt the group. % append an additional flag to each label, indicating the target of that post. 
To illustrate the reasoning for the schema, we show a series of example post templates and their corresponding labels in \textbf{Table \ref{tab:schema}}. Note that in the traditional labeling schema, as proposed in \citep{liu2019fuzzy} and widely followed in the MTL literature, a) any post that is toxic towards a specific group is considered non-toxic towards every other group (see rows 2 and 3); and b) any post that is non-toxic to a group is considered non-toxic towards every other group as well (see rows 5 and 6). Our conditional schema enables each demographic branch of the CondMTL model to conditionally filter out irrelevant examples (both toxic and non-toxic) for each group, and compute the loss over the relevant examples only.

\begin{table}[ht]
    \centering
    \vspace{-0.5em}
    \resizebox{0.65\linewidth}{!}{%
    \begin{tabular}{l|rr|rr} \toprule
        & \multicolumn{2}{c|}{TradMTL Label} & \multicolumn{2}{c}{CondMTL Label} \\
        Hypothetical Post & Men & Women & Men & Women \\ \midrule
        ``I hate everybody'' & T & T & T & T \\
        ``I hate men'' & T & {\color{red} \bf NT} & T & {\color{blue} \bf $\bullet$} \\
        ``I hate women'' & {\color{red} \bf NT} & T & {\color{blue} \bf $\bullet$} & T \\
        \hline
        ``I love everybody'' & NT & NT & NT & NT \\
        ``I love men'' & NT & {\color{red} \bf NT} & NT & {\color{blue} \bf $\bullet$} \\
        ``I love women'' & {\color{red} \bf NT} & NT & {\color{blue} \bf $\bullet$} & NT \\ \bottomrule
        \end{tabular}
        % I hate everybody & T & T & T & T & (T, M) & (T, F) \\
        % I hate men & T & T & {\color{red} \bf NT} & T & (T, M) & {\color{blue} \bf ($\bullet$, M)} \\
        % I hate women & T & {\color{red} \bf NT} & T & T & {\color{blue} \bf ($\bullet$, F)} & (T, F) \\
        % \hline
        % I love everybody & NT & NT & NT & NT & (NT, M) & (NT, F) \\
        % I love men & NT & NT & {\color{red} \bf NT} & NT & (NT, M) & {\color{blue} \bf ($\bullet$, M)} \\
        % I love women & NT & {\color{red} \bf NT} & NT & NT & {\color{blue} \bf ($\bullet$, F)} & (NT, F) \\ \bottomrule    
    }
    \caption{\small CondMTL group-specific labels \vs TradMTL labels for some hypothetical posts. T and NT denote toxic and non-toxic labels, % (Toxic) and NT (Non-Toxic) can be easily replaced with any suitable numbers by the practitioner, depending on the framework used. 
    % while M and W denote Men and Women groups. 
    The label ($t$) denotes the toxicity $t$ of a post towards a target group $k$. The {\color{blue}$\bullet$} indicates unknown toxicity wrt.\ the given group, whereas TradMTL methods erroneously assume such training examples are {\color{red}non-toxic}.}
    \vspace{-1.5em}
    \label{tab:schema}
\end{table}

\subsection{CondMTL Algorithm}

We follow the same architecture as TradMTL (shown in Fig. \ref{fig:arch-mtl}), but instead of standard weighted Binary Cross Entropy (wBCE), we use a Conditional weighted BCE loss function. wBCE is a variation of BCE that re-weights the error for the different classes proportional to their inverse frequency in the data \citep{lin2017focal}. This strategy is available in popular packages like SkLearn \citep{scikit-learn} and is useful to address class imbalance (\eg between toxic \vs non-toxic examples).
\vspace{-0.5em}
%\ma{what is "sub" in Algorithm 1? }\sg{updated algo}
% \ma{demographic flag in line 1 not shown}
\begin{algorithm}[htb]
	\caption{Conditional MTL Loss ($y_{\textrm{true}}, y_{\textrm{pred}}$)}
    \footnotesize
	\begin{algorithmic}[1]
	\BState \textbf{Input:} True Label $y_{\textrm{true}}= y$ \Comment{true label \wrt current branch}
	\BState \textbf{Input:} Predicted Label $y_{\textrm{pred}}=\hat{y}$ \Comment{Predicted probability of classifier}
	\BState \textbf{Input:} Class Weights $w_{\textrm{toxic}}, w_{\textrm{non-toxic}}$ \Comment{Assigned weights of classes} \vspace{-1em}
	\[\] {Select demographic relevant examples in current mini batch} %
	\State $y_{\textrm{true}}^k, y_{\textrm{pred}}^k = \{\}, \{\}$ \Comment{Empty lists to hold selected examples}
	\For {$i \in n$} \Comment{Loop over examples in current mini batch}
	    \If {$y \in k$} \Comment{current example is relevant to branch $k$}
	        \State $y_{\textrm{true}}^k=y_{\textrm{true}}^k \cup y$ \Comment{Append current label for consideration}
	        \State $y_{\textrm{pred}}^k=y_{\textrm{pred}}^k \cup \hat{y}$
	    \EndIf
	\EndFor \vspace{-1em}
	\[\] {Compute weighted BCE loss over relevant selected subset of examples}
	\State $err = wBCE(y_{\textrm{true}}^k, y_{\textrm{pred}}^k, w_{\textrm{toxic}}, w_{\textrm{non-toxic}})$
	\BState \textbf{Output}: Error for backpropagation $err$ 
	\end{algorithmic} \label{alg:condloss}
    % \vspace{-1em}
\end{algorithm}

% \ma{added generic explanation}
For a given MTL architecture, we can consider $K+1$ tasks: a generic one and $K$ group-specific ones. For example, if groups correspond to a simplified version of gender, with two possible group labels, then we would have three tasks: T1) given a post, is it toxic?; T2) given a post, is it toxic towards men? and T3) given a post, is it toxic towards women? All the examples in $\mathcal{D}$ are passed through the network, where the T1 branch learns a demographic-independent toxic \vs non-toxic representation over $N$ examples. While all $N$ examples and their labels get passed to the demographic-specific branches (T2 and T3) as well, CondMTL only allows back propagation for the relevant instances. For instance, only the $N_1$ examples of $\mathcal{D}_1$ that are targeted towards women demographics would be considered by the women branch. %Similarly, the women branch has access to all $N$ examples of $D$, but only calculates error over $N_2$ examples of $\mathcal{D}_2$ for T3.

% The architecture remaining the same\ma{the same to what? you haven't introduced the architecture yet}, all the examples in $\mathcal{D}$ are being passed through the network, where the All branch\ma{you haven't introduced the All branch} learns a demographic-independent toxic \vs non-toxic representation over $N$ examples. While all $N$ examples and their labels get passed to the demographic-specific branches as well, our CondMTL would only allow back propagation to happen on $N_1$ examples of $\mathcal{D}_1$ that are targeted towards men demographics only. Similarly, the women branch has access to all $N$ examples of $D$, but only calculates error over $N_2$ examples of $\mathcal{D}_2$.

% \vspace{-0.5em}

The Conditional Loss Function CondMTL is shown in \textbf{Alg. \ref{alg:condloss}}, which operates over each mini batch of examples to compute errors for backpropagation (\textit{steps 5-8}). It accepts two arguments, the true labels ($y_{\textrm{true}}= y$) and the predicted labels ($y_{\textrm{pred}}=\hat{y}$). Note that in our CondMTL loss, we are using the conditional label format as shown in Table \ref{tab:schema}, thereby $y_{\textrm{true}}$ is the label conditioned on the demographic flag. Iterating over each example (\textit{step 5}) in the mini batch, we only select relevant instances to that demographic branch based on the demographic flag $(k)$ (\textit{step 6}) and append the true and predicted labels to $y_{\textrm{true}}^k, y_{\textrm{pred}}^k$, respectively (\textit{step 7-8}). We also have the weights for each class ($w_{\textrm{toxic}}, w_{\textrm{non-toxic}}$), which are pre-computed during label generation. These weights can also be computed over each mini batch on the basis of the number of toxic \vs non-toxic examples in the selected subset $y_{\textrm{true}}^k$. We leave the choice of selecting weights up to the practitioner to account for class imbalance. Finally, we compute the weighted BCE loss on the selected relevant examples for backpropagation (\textit{step 9}).

When considering the template posts from \textbf{Table \ref{tab:schema}}, the TradMTL model with its labeling schema \citep{liu2019fuzzy} would correctly backpropagate its losses for the all, men, and women branches for the example \textit{I hate everybody}. Given that this post does target both the men and women groups and is toxic, the traditional label (T, T, T) is equivalent to the conditional label (T, T, T). However, the template post \textit{I hate men} reveals an issue with the traditional labeling schema and the subsequent information that a TradMTL model would learn; the traditional MTL model would backpropagate a misleading loss for the women branch due to the women label in the traditional label (T, T, NT) being marked as non-toxic (NT). The traditional MTL model would erroneously learn that a post which is toxic towards men is nontoxic if it were targeted at women, ultimately confusing the model. In contrast, the CondMTL model avoids backpropagating the loss which may confuse the model by examining the demographic flag corresponding to the label (T, T, $\bullet$) and using it to compute the loss only for the men branch.%\st{, while ignoring it when computing the loss of the women branch. Similar logic would follow for the \textit{I hate women} post, which is explicitly targeted toward women, hence should not be considered to be a part of the men branch.}

\vspace{-0.5em}
\section{Experimental Details} \label{sec:data}

We describe the dataset used for validation along with a stacked single task model baseline and other MTL models for comparison with CondMTL. Implementation details regarding setup, loss curves, and class balancing strategy are enumerated in \textbf{Appendix \ref{app:setup}}.
\vspace{-0.5em}

\subsection{Data}

\noindent \textbf{Target Identity Dataset.} To assess differential subgroup validity in toxic language detection, we focus on toxicity and gender~\citep{rubin2020fragile,vogels2021}. 
% \ma{what does "fair protection of different groups" mean?}, 
We use the {\em Civil Comments} \citep{borkan2019nuanced} portion of {\em Wilds}
\citep{koh2021wilds}. The dataset has 48,588 training posts labeled as Toxic or non-Toxic. Each post has an explicit annotation for the demographics \ie gender groups of the target entity, with probability scores about the annotator consensus. We select posts where more than 50\% of annotators agreed on the gender of the target. 
We include only women (W) and men (M)
% \ma{we should talk about women and men since we are referring to gender}
genders, to construct a simplified binary sensitive attribute for our experiments. However, we emphasize that this is a simplification and acknowledge the non-binary nature of gender. Moreover, we note that the reliance on annotators to identify the gender of the target may contain errors.
% \ma{commented out saying that we did it for convenience because it may lead people to think the method doesn't work well for non-binary attributes, and it was also somewhat confusing after saying the method was agnostic }%All methods used in this work are agnostic to the type of sensitive attribute.%, and our inclusion of only two genders merely reflects a convenient way to assess the capabilities of our proposed method in regard to maximizing accuracy across a binary sensitive attribute.

We consider posts where either group (women: 22,149 and men: 15,305) or both groups (both: 11,134) are targeted. %For posts with both targets present, we duplicate them for the men and women parts. 
% \ma{commented out the duplication, since it is confusing and we already explained the labeling schema}
\textbf{Table \ref{tab:wilds-stats}} shows the distribution of targets and labels in the dataset. We use the same procedure for both the train and test splits from the dataset. We observe roughly a 15\%-85\% split between toxic \vs non-toxic labels across all three branches for both the train and test splits. %\st{This indicates the imbalance in the labels where toxic speech is the smaller class; \ie in the dataset the prevalence of toxic examples are scarce.}

\begin{table}[ht]
    \centering
    \vspace{-0.5em}
    \resizebox{0.85\linewidth}{!}{%
    \begin{tabular}{l|rrr|rrr}\toprule
        & \multicolumn{3}{c|}{Train Split} &  \multicolumn{3}{c}{Test Split} \\ \midrule
		Branch & Toxic & Non-Toxic & Total & Toxic & Non-Toxic & Total\\ \midrule
		All & 7,099 (14\%) & 41,489 (86\%) & 48,588 & 3,350 (15\%) & 19,236 (85\%) & 22,586\\
		Men (M) & 3,940 (15\%) & 22,499 (85\%) & 26,439 & 1,920 (15\%) & 10,694 (85\%) & 12,614\\
		Women (W) & 4,560 (14\%) & 28,723 (86\%) & 33,283 & 2,068 (14\%) & 12,964 (86\%) & 15,032\\ \bottomrule
	\end{tabular}
	}
% 	\vspace{-1em}
	\caption{\small Statistics of the Wilds \citep{koh2021wilds} dataset. We consider the binary sensitive target gender as men \vs women. The all branch contains all data points, while men and women branches contain the data points in which posts target men or women groups, respectively.}
	\vspace{-1.5em}
    \label{tab:wilds-stats}
\end{table}

\subsection{Baselines}

\begin{figure}[ht]
    \centering
    % \vspace{-0.5em}
    \includegraphics[width=\linewidth]{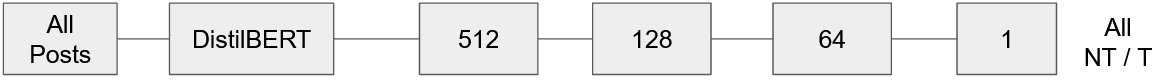}
    \vspace{-1.5em}
    \caption{\small Architecture for Single Task, where all posts are passed though a neural network and get classified as toxic \vs non-toxic.}
    % \vspace{-0.5em}
    \label{fig:arch-stl}
\end{figure}

For a single task (\textbf{STL}) baseline, we use a DistilBERT \citep{sanh2019distilbert} representation layer to extract numerical features from posts. This is followed by layers of dense neuron connections with \textit{relu} activation and added biases, ending in a classification node with \textit{sigmoid} activation with 0.5 classification threshold (\textbf{Fig. \ref{fig:arch-stl}}). For our experiments, we freeze the weights of the DistilBERT \citep{sanh2019distilbert} representation layer. The only trainable parameters in the models are the dense neuron units that follow the DistilBERT layer until the output branch. One can replace the DistilBERT layer with any other advanced feature representation without altering the rest of the model. 

\begin{figure}[ht]
    \centering
    % \vspace{-0.5em}
    \includegraphics[width=\linewidth]{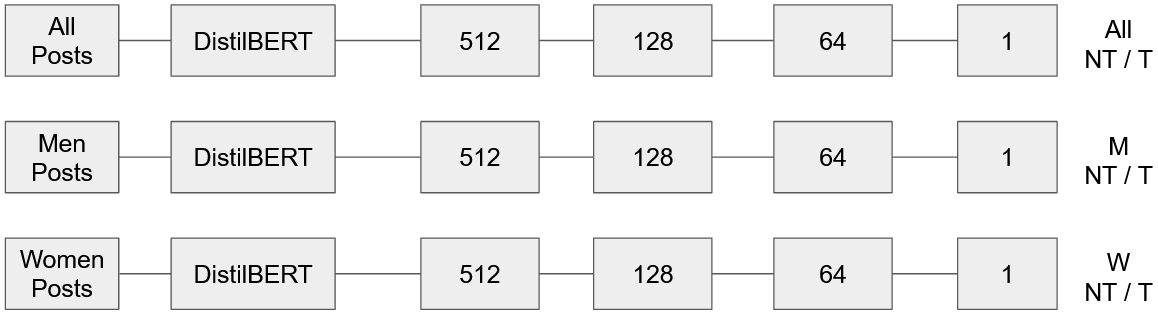}
    \vspace{-1.5em}
    \caption{\small Architecture for stacked STL contains three independent single task models, one for each portion of the data.}
    % \vspace{-0.5em}
    \label{fig:arch-mstl}
\end{figure}

\noindent \textbf{Stacked STL} model (\textbf{Fig. \ref{fig:arch-mstl}}) contains independent classifiers  for each demographics, distinguishing toxic \vs non-toxic. For the Wilds dataset, we construct All, Men, and Women classifiers resulting in $3\times$ the trainable parameters of one Single task classifier. 

\begin{figure}[ht]
    \centering
    % \vspace{-0.5em}
    \includegraphics[width=\linewidth]{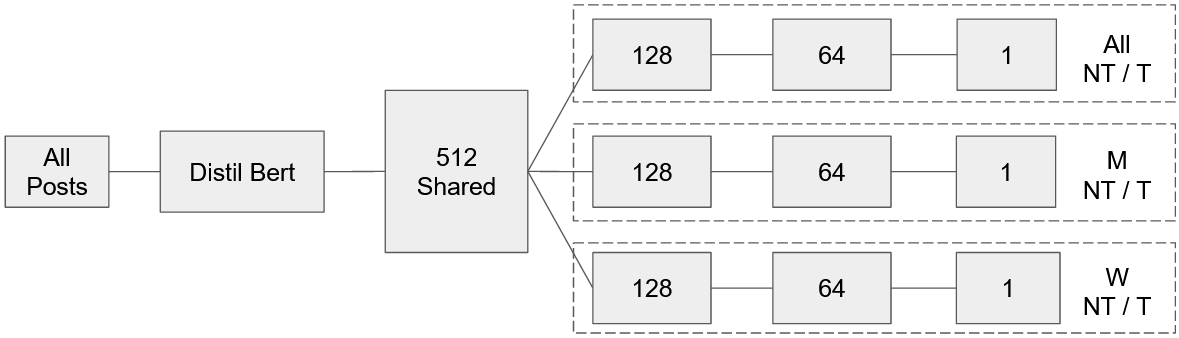}
    \vspace{-2.0em}
    \caption{\small Architecture for TradMTL and CondMTL, where the 512 dense neurons are shared across all three tasks while maintaining independent task specific layers (mark by dashed boxes).}
    % \vspace{-0.5em}
    \label{fig:arch-mtl}
\end{figure}

\noindent \textbf{Traditional Multi Task (TradMTL)} model (\textbf{Fig. \ref{fig:arch-mtl}}) contains a shared layer of 512 dense neurons across all the tasks, while the individual task-specific layers (enclosed in dashed boxes) have dense connections of 128, 64 and 1 each, following the architecture of the STL model. The shared layer is responsible for learning a representation that is common across all tasks, while the task specific layers learn representations specific to their own tasks for differentiating between toxic \vs non-toxic posts.

\begin{figure}[ht]
    \centering
    % \vspace{-0.5em}
    \includegraphics[width=\linewidth]{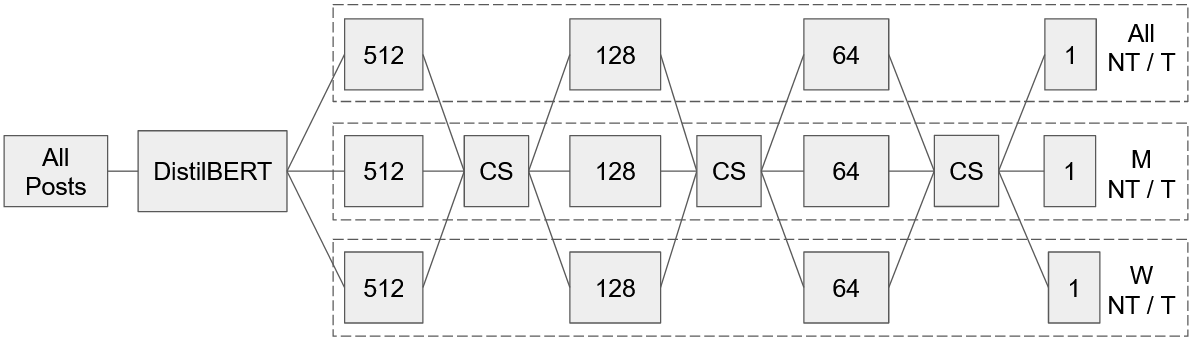}
    \vspace{-2.0em}
    \caption{\small Architecture for CSMTL, which is  replica of the Stacked STL model, with cross stitch (CS) units between each dense layers, allowing them to share weights across tasks for task similarity.}
    \vspace{-1.5em}
    \label{fig:arch-cstl}
\end{figure}

\noindent \textbf{Cross Stitch Multi Task (CSMTL)} model (\textbf{Fig. \ref{fig:arch-cstl}}) is similar to the stacked STL model (Fig. \ref{fig:arch-mstl}) with Cross Stitch (CS) units \citep{misra2016cross} placed between each dense layer. The CS layer is a $K \times K$ weight matrix, initialized as Identity $I_K$. The intuition is that if the $K$ tasks are independent, then the identity holds even after training with backpropagation. If the tasks are correlated, then the CS matrix at each layer would deviate from identity and learn some common correlation structure across similar tasks. However, both theoretically and empirically, the CS structure does not always improve performance, while taking up more than $K \times$ trainable parameters. We choose this framework for comparison, as it is one of the most widely used ones in the MTL literature.

\subsection{Inference}
All models discussed (STL, TradMTL, CSMTL, and CondMTL) assume group labels at inference time. Each post in the {\em Civil Comments} \citep{borkan2019nuanced} portion of {\em Wilds}
\citep{koh2021wilds} is human-annotated for the demographics, which includes the gender group of the target entity of each post. We acknowledge that human annotations may contain errors.

\section{Results} \label{sec:res}

We present the performance of CondMTL \wrt the stacked STL baseline and other MTL variants in terms of trainable parameters and training runtime. We examine post hoc classification performance and recall disparities and compare confusion matrices for all of the models. We also present analysis of our CondMTL and CSMTL in terms of label contamination and contamination of weights.

\subsection{Architecture and Runtime}

\textbf{Table \ref{tab:wilds-arch}} shows trainable parameters for the baseline models and for CondMTL. 
% \ma{check placement of this sentence I moved}
Results shown are only over one run. For mean and variance reports over multiple runs, please refer to \textbf{Appendix \ref{app:variance}}. %Note that the DistilBERT representation layer weights are frozen, and only the dense layers contain trainable parameters. Each dense neuron has \textit{relu} activation with bias enabled. The output neuron has sigmoid activation, with a threshold of 0.5 for classifying toxic (0) \vs non-toxic (1). 
%Examining the parameter size differences between the models parameterized in Table \ref{tab:wilds-arch} suggests that choosing either the TradMTL or CondMTL models may be preferable to the Stacked STL or CSMTL models, because the size differences result in faster training times and reduced computational load which are important factors in model deployment. Current results shown are only over one run. For mean and variance reports over multiple runs, please refer to \textbf{Appendix \ref{app:variance}}.

% \begin{table}[ht]
% \centering
%     \vspace{-0.5em}
%     % \resizebox{\linewidth}{!}{%
% 	\begin{tabular}{l|rr} \toprule
% 	    Model type & \# Params ($\times$ Single Task) & Time (s) \\ \midrule
%         Stacked STL & 1,403,139 ($3 \times$) & 2,400 \\ \midrule
% 		Trad MTL &  615,683 ($1.31 \times$) & 2,200 \\
% 		CS MTL \citep{misra2016cross} & 1,403,166 ($3.05 \times$) & 2,600 \\ \midrule
% 		Cond MTL (Ours) & 615,683 ($1.31 \times$) & 2,050 \\ \bottomrule
% 	\end{tabular}
% % 	}
% % 	\vspace{-0.5em}
% 	\caption{\small Architecture Parameters for Wilds dataset with three tasks. The DistilBERT representation is frozen and the dense layers are trainable, with each STL model having 467,713 trainable parameters. We also report the training time for the models over 10 epochs.}
% 	\label{tab:wilds-arch}
%     \vspace{-1.5em}
% \end{table}

\begin{table}[ht]
\centering
    % \vspace{-0.5em}
    \resizebox{0.7\linewidth}{!}{%
	\begin{tabular}{l|rr|rr} \toprule
	    Model type & \# Params & $\Delta$ & Time(s) & $\Delta$ \\ \midrule
        Stacked STL (3 models) & 1,403,139 & - & 7,200 & -\\ \midrule
		CSMTL \citep{misra2016cross} & 1,403,166 & +0\% & 2,600 & -64\%\\ 
		TradMTL &  615,683 & -56\% & 2,200 & -69\%\\
%		\midrule
		CondMTL (Ours) & 615,683 & -56\% & 2,050 & -72\%\\ \bottomrule
	\end{tabular}
  	}
% 	\vspace{-0.5em}
	\caption{\small Space (parameter size) and training time (seconds for 10 epochs) required by STL \vs MTL models on the Wilds dataset. The DistilBERT representation is frozen and the dense layers are trainable, with each STL model having 467,713 trainable parameters. For the 3 tasks considered, we assume 3 different STL models and report space and time summed over all 3. We then report \% space and time reduction achieved by MTL models \vs this baseline of 3 STL models.	}
    \vspace{-1.5em}
    \label{tab:wilds-arch}
\end{table}

The single task model (Fig. \ref{fig:arch-stl}) has 467,713 trainable parameters, hence the stacked STL (Fig. \ref{fig:arch-mstl}) operating on the All, Men, and Women portions of the data has $3\times$ or $1,403,139$ trainable parameters. We report space reduction achieved by MTL models \vs this reference of 3 STL models. The TradMTL and CondMTL models (Fig. \ref{fig:arch-mtl}) have the same architecture but different labeling schema and loss functions. They have a shared 512 unit layer representation and three task specific branches which collectively have 56\% fewer trainable parameters when compared to the stacked STL model. The CSMTL model (Fig. \ref{fig:arch-cstl}) is a replica of the Stacked Single Task model with cross stitch (CS) units between each of the dense layers. It has 27 ($\sim +0\%$) more trainable parameters when compared to the Stacked STL model due to the extra connections from the CS units. In terms of training and further deployment, the traditional and conditional MTL models are preferable due to significantly reduced model size even when dealing with multiple tasks (three in this case). One can observe that the trainable parameters in Cross Stitch networks scale linearly \wrt number of tasks, causing memory stagnation. This issue has been raised and studied in \citep{strezoski2019many}. 

We report the training runtime in Table \ref{tab:wilds-arch} \wrt 10 epochs. Both TradMTL and CondMTL models have the same number of trainable parameters. However the CondMTL model only trains over a subset of the data in its men and women branches, which reduces runtime. Empirically, we observe a reduction of 72\% in CondMTL \vs 69\% in TradMTL. The stacked STL and CSMTL models take longer to train, since they roughly have the same number of trainable parameters. However, the CSMTL model operates on the three branches in a single model rather than three independent models, resulting in a lower GPU pipeline load and a $64\%$ reduction in time.

\subsection{Performance Measures}

\begin{figure*}[ht]
    \centering
    \begin{subfigure}{0.49\linewidth}
        \includegraphics[width=\linewidth]{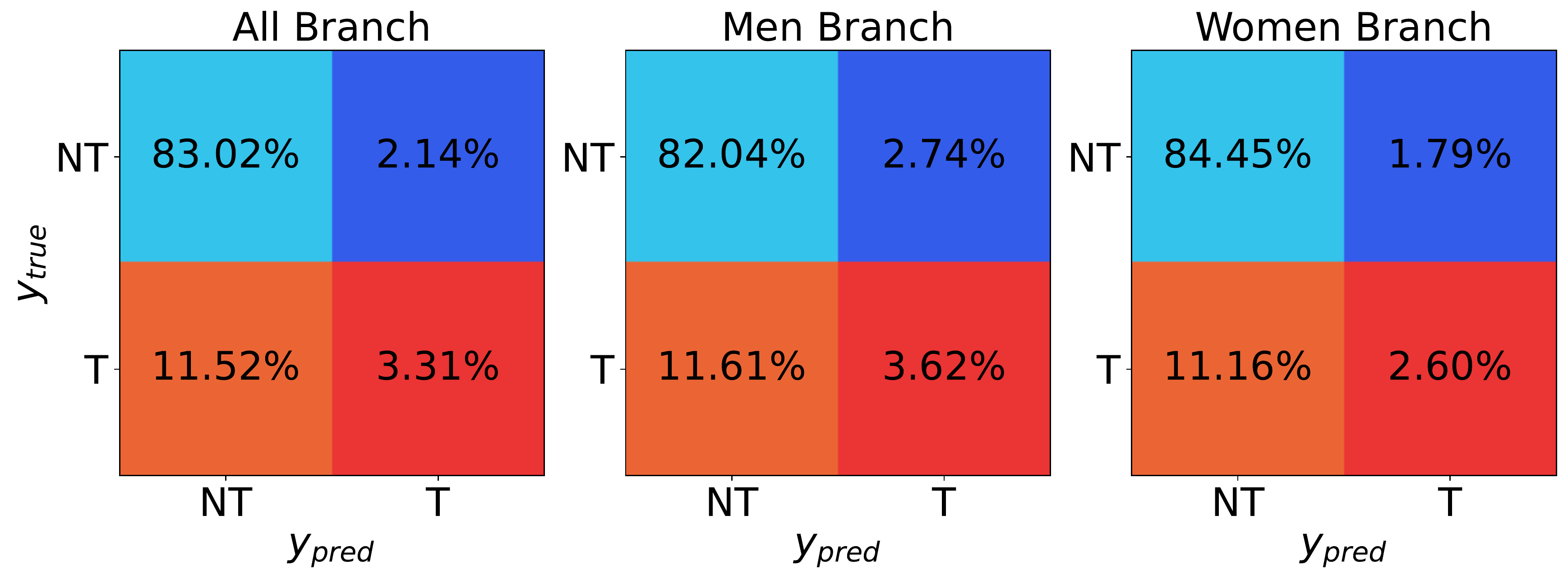}
        \vspace{-1.5em}
        \caption{\small Confusion matrices of 3 branches for Stacked STL.}
    \end{subfigure}
    % \quad
    \begin{subfigure}{0.49\linewidth}
        \vspace{1em}
        \includegraphics[width=\linewidth]{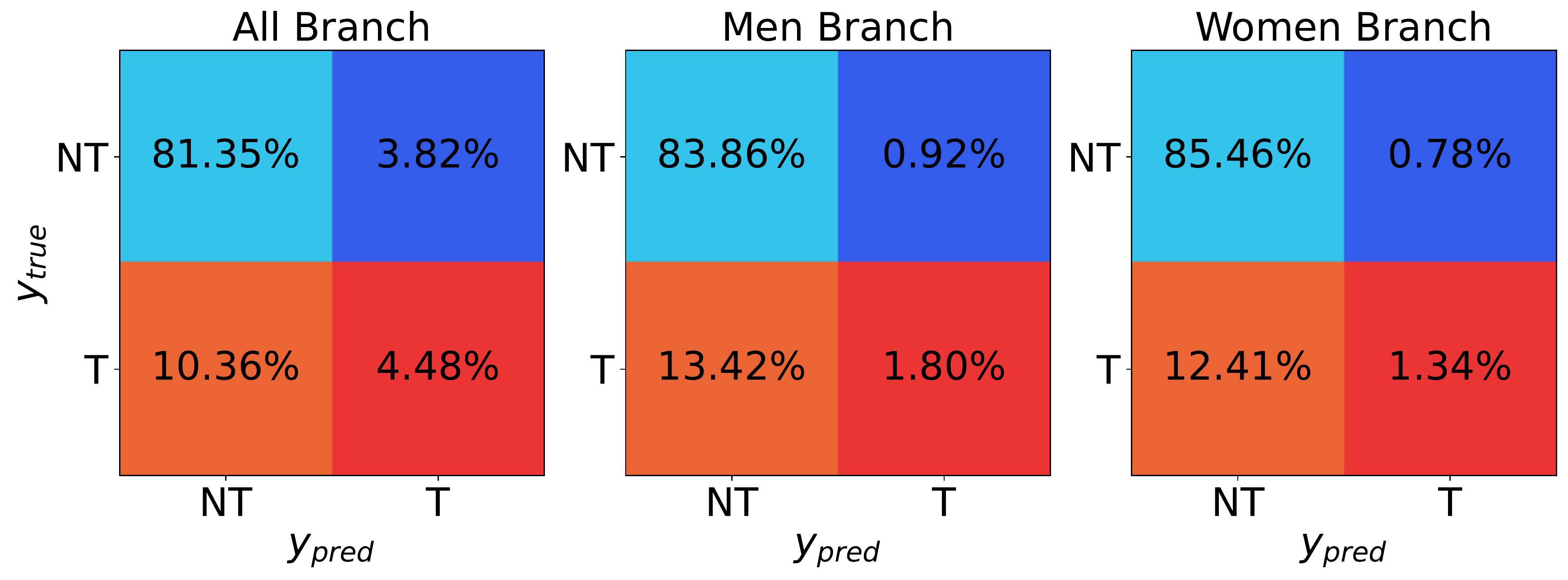}
        \vspace{-1.5em}
        \caption{\small Confusion matrices of 3 branches for Traditional MTL.}
    \end{subfigure}
    \begin{subfigure}{0.49\linewidth}
        \vspace{1.0em}
        \includegraphics[width=\linewidth]{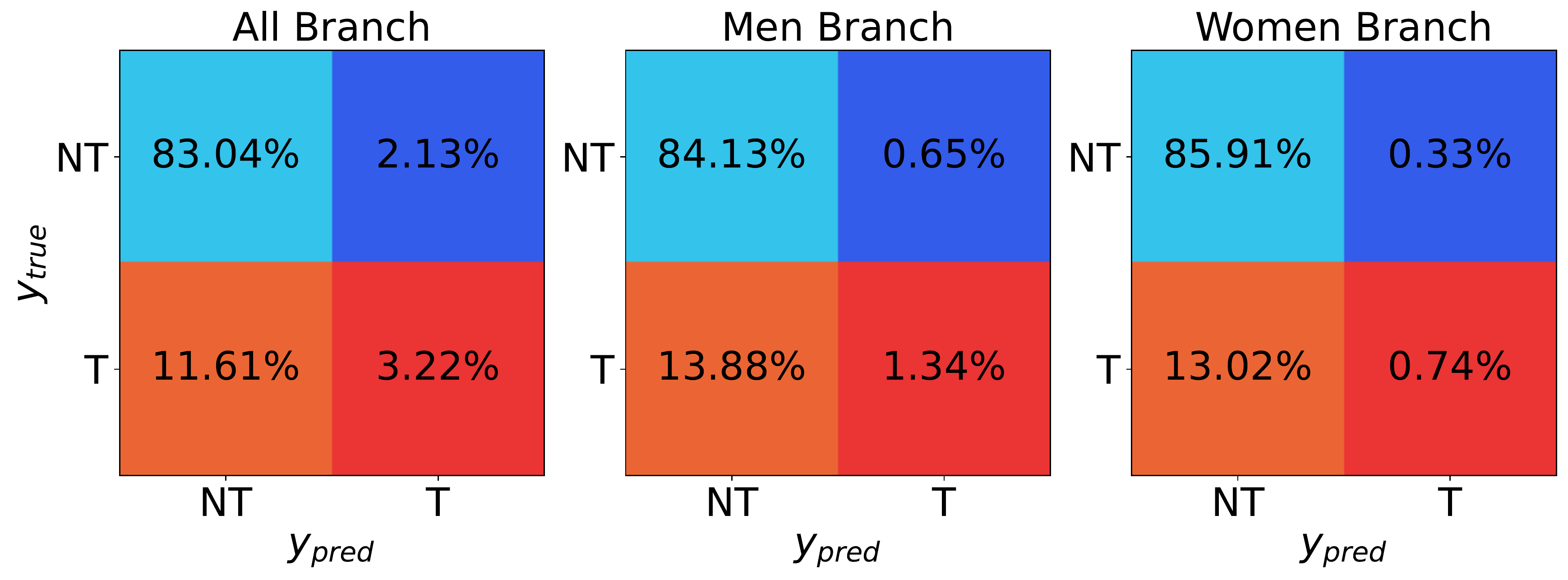}
        \vspace{-1.5em}
        \caption{\small Confusion matrices of 3 branches for Cross Stitch MTL.}
    \end{subfigure}
    % \quad
    \begin{subfigure}{0.49\linewidth}
        \vspace{1.0em}
        \includegraphics[width=\linewidth]{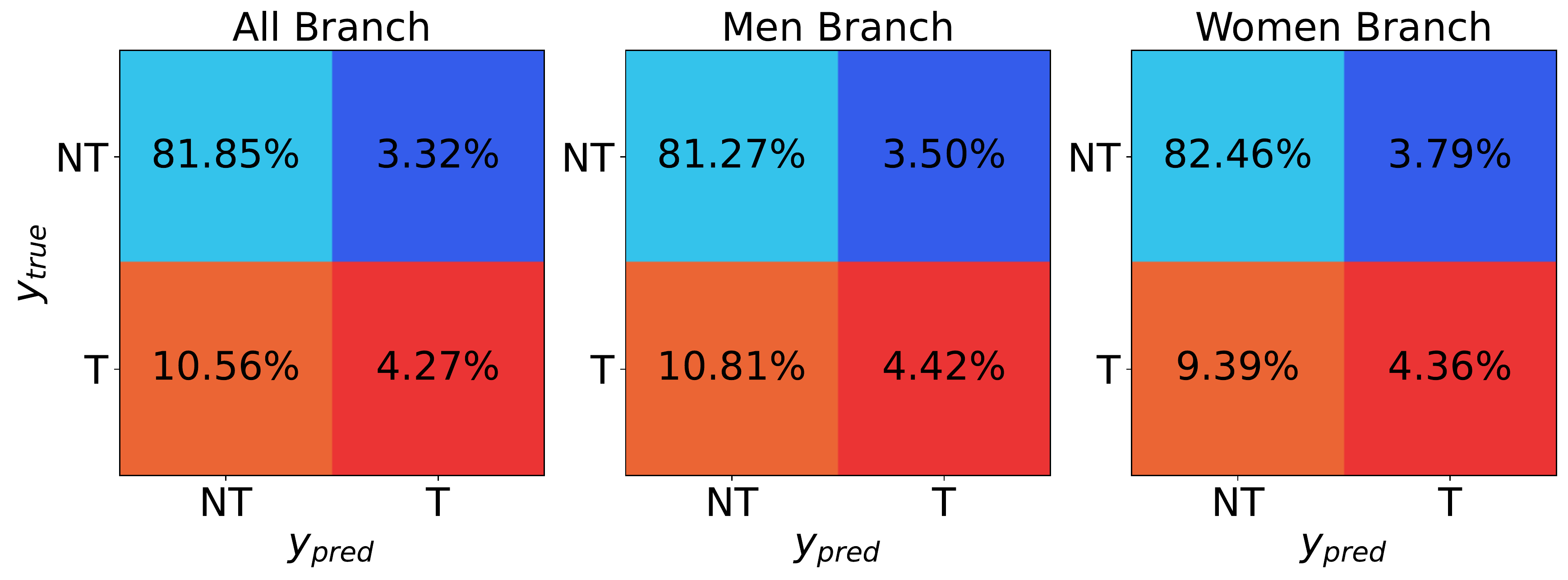}
        \vspace{-1.5em}
        \caption{\small Confusion matrices of 3 branches for Conditional MTL.}
    \end{subfigure}
    % \vspace{-1.0em}
    \caption{\small Confusion matrices of 3 tasks (columns) for different models (rows). Values are shown as percentages in each block \wrt the number of instances relevant to that branch. The performance of the models are comparable in the \emph{All} branch, since they train over the full dataset. CondMTL performs significantly better in the demographic specific \emph{men} and \emph{women} branches, due to training over group relevant examples only. This can be observed by the higher numbers in the {\color{red} red} boxes of these branches for CondMTL compared to the other MTL variants. The toxic class is the smaller class, %\ie with fewer occurrences, 
    hence a good toxicity detection model should be able to correctly identify as many of them as possible. A good model should also have fewer misses (toxic posts identified as non-toxic), indicated by the {\color{orange} orange} boxes, where also CondMTL performs best.}
    \vspace{-1.0em}
    \label{fig:conf_mat}
\end{figure*}

We show the performance comparison of the models on the Wilds-Civil Comments \citep{koh2021wilds} test dataset. The Accuracy numbers in \textbf{Appendix \ref{app:variance}: Table \ref{tab:wilds-acc}} indicate that all of the models roughly perform the same in terms of overall accuracy \wrt the Stacked STL ($\sim 86\%$). Since the dataset is imbalanced, with the non-toxic class encompassing 85\% of the labels, a model which trivially predicts all testing posts as non-toxic would also achieve a roughly 85\% accuracy score. Given that all models perform approximately the same as this trivial baseline, we need to consider other metrics to more holistically evaluate model performance. 
% \ma{I am confused by  the placement of parameter discussion here, this seems to belong in 6.1 (architecture), and we already discussed it, so I'm commenting out and moving up}%Examining the parameter size differences between the models parameterized in Table \ref{tab:wilds-arch} suggests that choosing either the TradMTL or CondMTL models may be preferable to the Stacked STL or CSMTL models, because the size differences result in faster training times and reduced computational load which are important factors in model deployment. Results shown are only over one run. For mean and variance reports over multiple runs, please refer to \textbf{Appendix \ref{app:variance}}.

\begin{table}[ht]
    \centering
    \resizebox{0.7\linewidth}{!}{%
    \begin{tabular}{l|l|rr|rr|rr} \toprule
        & & \multicolumn{2}{c|}{All} & \multicolumn{2}{c|}{Men} & \multicolumn{2}{c}{Women} \\
        & & NT & T & NT & T & NT & T \\ \midrule    
        \multirow{4}{*}{Recall} & Stacked STL & 96.9 & 25.2 & 96.8 & 23.8 & 97.1 & 23.6 \\
        & CSMTL & 97.2 & 23.6 & 98.8 & 13.3 & 99.7 & 4.6 \\
        & TradMTL & 94.4 & 20.1 & 95.8 & 4.2 & 95.6 & 2.8 \\
        & CondMTL (Ours) & 96.1 & 29.0 & 95.9 & \textbf{28.7} & 95.1 & \textbf{31.2} \\ \midrule
        \multirow{4}{*}{F1} & Stacked STL & 92.3 & 35.3 & 92.0 & 33.5 & 92.8 & 33.3 \\
        & CSMTL & 92.3 & 33.8 & 92.2 & 22.2 & 92.8 & 8.7 \\
        & TradMTL & 92.1 & 29.8 & 91.9 & 7.9 & 92.6 & 5.4 \\ 
        & CondMTL (Ours) & 92.2 & 38.3 & 92.9 & \textbf{37.9} & 93.6 & \textbf{39.5} \\ \midrule
        \multirow{4}{*}{Precision} & Stacked STL & 88.2 & 58.6 & 87.6 & 56.9 & 88.9 & 56.4 \\
        & CSMTL & 88.0 & 59.3 & 86.4 & \textbf{67.0} & 86.8 & \textbf{69.1} \\
        & TradMTL & 87.5 & 54.4 & 85.3 & 47.7 & 86.6 & 46.7 \\
        & CondMTL (Ours) & 88.6 & 56.1 & 88.2 & 55.9 & 89.7 & 53.7 \\ \bottomrule
    \end{tabular}
    }
    \caption{\small Statistic Comparison between different methods based on internal stats: Recall, F1 and Precision. Numbers are bolded only when they are significantly better than the other models. For a toxic language detection task, \textit{Recall} is of prime importance over the smaller toxic class, since we want to detect as many of the toxic posts as possible in deployment. We observe that CondMTL achieves significantly better recall values for both groups on the toxic labels.}
    \vspace{-1.5em}
    \label{tab:wilds-results}
\end{table}

In order to identify the discrepancies between the models, we compare the Recall, F1 and Precision scores in \textbf{Table \ref{tab:wilds-results}}. Since the dataset is imbalanced with roughly a 85\%-15\% split between the non-toxic \vs toxic labels, we observe the bias of the classifier towards detection of non-toxic examples in spite of class re-weighting during model training.
%Recall is the most important metric to consider for a TL detection model since it demonstrates what proportion of the toxic posts the model detects. However, precision must be considered as well, since a model can trivially attain high recall values for the toxic class by predicting all test examples to be toxic. Precision reflects the model's accuracy in its predictions, as it is the proportion of the number of correct predictions for a class divided by the number of predictions the model makes for the class. Meanwhile, a model can attain high precision by only making predictions on examples for which it is very confident in (toxic examples which are very clearly toxic) while incorrectly labeling less toxic examples as non-toxic. Thus, while we would like a model which is precise in its predictions, considering both recall and precision in tandem are necessary to appropriately evaluate classification models. Subsequently we present the recall, precision, and F1 score (which combines the recall and precision as a harmonic mean) for all of the models. 
For the non-toxic (NT) class, all of the post hoc measures are roughly equivalent for the all branch ($96\%$ for Recall, $92\%$ for F1, and $88\%$ for Precision) across the compared models. Similar behavior is observed over the men and women branches.

We observe in Table \ref{tab:wilds-results} that CondMTL achieves better recall values \wrt baselines over the smaller \ie toxic class. %, since CondMTL ensures that the demographic group branches learn a more accurate understanding of toxicity.
CondMTL produces recall values of $(29\%, 31\%)$ for the men and women branches respectively, showing marked improvement over CSMTL and TradMTL, which produce recall values of $(13\%, 5\%)$ and $(4\%, 3\%)$, and also outperforming Stacked STL $(24\%,24\%)$. The superior performance of CondMTL in terms of recall, likely driven by its more accurate understanding of toxicity at a group-specific level, is crucial in the context of automated toxicity detection, where we would like to ensure that toxic posts are not mislabeled as non-toxic (misses), as such errors could disproportionately affect marginalized demographic groups. For instance, women are disproportionately affected by stalking and by sexualized forms of abuse~\citep{vogels2021}. %In the context of gender, for example, online harassment targeting women often involves more serious violations %For this reason, achieving a higher recall (less false negatives) is typically considered to be more important than precision in evaluating toxicity detection tasks. 

In terms of precision, CSMTL performs the best ($67\%,69\%$) compared to TradMTL ($48\%,47\%$) and CondMTL ($56\%,54\%$). CSMTL's higher precision number suggests that it is more reserved when predicting a test example to be toxic, which results in less false alarms (\ie a non-toxic post that is erroneously flagged). 

F1 %\st{, being a harmonic mean of precision and recall,}
provides a joint view of both precision and recall. In terms of F1, we observe that CondMTL ($38\%,40\%$) provides the best results, outperforming CSMTL ($22\%,9\%$), TradMTL ($8\%,5\%$) and Stacked STL $(34\%,33\%)$. This is because CondMTL's recall values are a scale apart compared to the other models.%\st{, in particular the other MTL variants.}

Although our CondMTL model is not optimized over any strict differentiable measure of fairness, we post hoc observe that it has a low false negative error rate balance \ie improved equal opportunity \citep{hardt2016equality}. Mathematically, a classifier with equal false negative rate (FNR) will also have equal true positive rate (TPR) or recall. We have shown that CondMTL achieves much better recall values compared to other MTL variants. \textbf{Table \ref{tab:eo}} shows the post-hoc measured Equal Opportunity (EO) gap across both groups for the models. All models except for CSMTL ($9.0$) produce low EO gaps. %, compared to $1.4$ and $2.5$ for TradMTL and CondMTL respectively. 
Although having a lower EO gap value is ideal, it is necessary to evaluate the EO gap values of the different models with the context of their recall values. Thus, while TradMTL has the lowest EO gap value (1.4) among the MTL variants, given its poor recall values %\st{($4.2$ and $2.8$ for the men and women branches)} 
this model is unlikely to be desirable in practice, whereas CondMTL produces a low EO gap of 2.5 while maintaining higher recall values.

\begin{table}[ht]
    \centering
    \vspace{1em}
    \resizebox{0.65\linewidth}{!}{%
    \begin{tabular}{l|ccc} \toprule
        Model & Recall (Men) & Recall (Women) & EO Gap \\ \midrule
        Stacked STL & 23.8 & 23.6 & 0.2 \\
        CSMTL & 13.6 & 4.6 & 9.0 \\
        TradMTL & 4.2 & 2.8 & 1.4 \\
        CondMTL (Ours) & 28.7 & 31.2 & 2.5 \\ \bottomrule
    \end{tabular}
    }
    \caption{\small Recall per group and Equal Opportunity (EO) gap measured as the absolute difference between recall values over the groups. For recall, higher values are better. For EO, lower values are better.}
    \vspace{-1.5em}
    \label{tab:eo}
\end{table}

Comparing the confusion matrices of the three branches of the MTL models in \textbf{Fig. \ref{fig:conf_mat}} reveals that CondMTL performs better in the demographic group branches (men and women) for the smaller toxic class. All models perform fairly well when classifying the nontoxic examples \ie the {\color{cyan} cyan} sections. Non-toxic posts that are erroneously flagged as toxic (\ie false alarms) are shown in the {\color{blue}blue} sections. On the other hand, TradMTL and CSMTL both struggle to correctly identify toxic examples and instead classify a greater portion of the toxic test examples as nontoxic (\ie misses). These misses correspond to the {\color{orange}orange} sections of the confusion matrices. Comparing the {\color{red}red} sections of the model confusion matrices reveals that CondMTL correctly classifies a greater proportion of the smaller toxic class. %In the context of toxicity detection, 
Given that non-toxic language is more common, %\st{the data distribution of toxicity datasets is often skewed towards non-toxic (as most language in the real world is non-toxic)} most language in the real world is non-toxic,
CondMTL's ability to capture a greater proportion of the toxic posts would be valuable in a deployed %automated 
toxicity detection model.

\vspace{-0.5em}
\subsection{Analysis of Conditional MTL} \label{sec:ana}

We make two propositions based on the theoretical working and empirical analysis of the CondMTL and CSMTL networks. Furthermore, we verify our stated propositions for CondMTL and CSMTL through simple and verifiable benchmarking cases in \textbf{Appendix \ref{app:benchmark}}, both on a regression and classification task.

\begin{proposition}
    Our proposed Conditional MTL does not allow contamination of weights across shared task layers and learns only over the group specific distribution for each demographic branch. 
\end{proposition}

The CondMTL architecture (Fig. \ref{fig:arch-mtl}) is an exact copy of TradMTL with the distinction of the updated loss function and labeling schema. Since the task specific layers (indicated by dashed boxes) do not interact with each other, each loss function is strictly guided by the examples that are relevant to its own branch. Assuming that the data distribution \wrt two groups $\mathcal{D}_1$ and $\mathcal{D}_2$ are  independent of each other, each branch learns a representation of their own dataset and does not take into account group irrelevant examples. The CondMTL loss (Alg. \ref{alg:condloss}) computes the loss over each group-specific distribution (Eq. \ref{eq:condmtl}), thereby avoiding label contamination.
\begin{align*}
    err_{all} &= wBCE([y_{true}]_{\mathcal{D}}, [y_{pred}]_{\mathcal{D}})\\
    err_{men} &= wBCE([y_{true}]_{\mathcal{D}_1}, [y_{pred}]_{\mathcal{D}_1}) \\
    err_{women} &= wBCE([y_{true}]_{\mathcal{D}_2}, [y_{pred}]_{\mathcal{D}_2}) \numberthis{} \label{eq:condmtl}
\end{align*}

\begin{proposition}
    Cross Stitch MTL \citep{misra2016cross} allows contamination of weights across shared task layers.
\end{proposition}

The CS unit (Fig. \ref{fig:arch-cstl}) is initialized with an identity structure, where the number of tasks dictates the size of the matrix. For illustration, let us consider two tasks for $CS \in I_2$. We find the following two flaws \wrt the logic of CS units: a) if the two tasks are truly independent, then the CS unit should not deviate from identity; and b) even when two tasks are correlated, allowing deviation from identity, the CS unit should still be a symmetric matrix since two tasks talking to each other are symmetrically equivalent. However, such constraints are not present in the implementation of the CS units, which causes them to learn arbitrary weights during model training. The weights become cross-contaminated across tasks.

To verify illustration 2, we also show the final weights of the CS units in our Wilds dataset training. One can observe that the symmetric property of CS units is violated. Note that due to the \textit{same same but different} nature of group-targeted toxicity, they share some commonality \ie they are not fully independent of each other, which would cause the CS matrix to deviate from identity. However, since tasks talking amongst each other should be symmetrical in nature, we would expect the updated CS matrix to hold the symmetric property. The values reported in Eq. \ref{eq:cs_wilds} are \wrt Fig. \ref{fig:arch-cstl} where we have three CS matrices. These three CS matrices show clear deviation from symmetry in the off-diagonal elements.

\resizebox{.85\linewidth}{!}{
    \begin{minipage}{\linewidth}
        \begin{align*}
        CS_1 &= \begin{bmatrix*}[r]
        1.00 & -3.69e-3 & -1.46e-4 \\
        1.53e-3 & 1.00 & 2.53e-3 \\
        -4.28e-3 & -1.04e-3 & 1.01\\
        \end{bmatrix*}\\
        CS_2 &= \begin{bmatrix*}[r]
        1.00 & 2.49e-2 & -2.29e-2 \\
        1.05e-2 & 1.01 & 2.99e-3 \\
        -1.03e-2 & 4.74e-3 & 1.01 \\
        \end{bmatrix*}\\
        CS_3 &= \begin{bmatrix*}[r]
        1.00 & 1.29e-2 & 5.13e-2 \\
        3.53e-2 & 1.01 & 2.19e-2 \\
        7.74e-5 & 9.51e-3 & 1.01 \\
        \end{bmatrix*} \numberthis{} \label{eq:cs_wilds}
        \end{align*}
    \end{minipage}
}

\section{Discussion and Future Work} \label{sec:disc}

% We perform comparative error analysis in \textbf{Table \ref{tab:disc}}, showing test examples which demonstrate the improvement of CondMTL \wrt TradMTL and CSMTL. The predicted labels for both TradMTL and CSMTL on all of the posts highlights the effects of label contamination on the demographic-specific task layers. As a result of the misleading labels discussed in Section \ref{sec:contam}, the TradMTL and CSMTL models have learned to mostly label examples as non-toxic (NT). When examining post 3 in Table \ref{tab:disc}, we can observe that it is both toxic and directed at the men demographic only; however, had post 3 been in the training set, it would have erroneously taught the women branch of the baseline MTL models that the post was nontoxic. Similarly, post 6 (toxic and targeted towards women only) would have contaminated the weights of the men branch of the baseline MTL models by skewing it to make more nontoxic predictions. While this will result in high accuracy scores for the models because the majority of the Wilds dataset is nontoxic, the baseline MTL models will subsequently acquire a poor understanding of demographic-targeted toxicity. Conversely, CondMTL ensures that the demographic group branches learn a more accurate understanding of toxicity and correctly label toxic posts as toxic. CondMTL correctly predicts that the posts in Table \ref{tab:disc} are toxic in the demographic-specific branches, and these predicted labels reflect the CondMTL model's better understanding of targeted TL. 

\textbf{Effect of Label Contamination.} As a result of the misleading labeling schema discussed in Section \ref{sec:contam}, the TradMTL and CSMTL models learn to mostly label examples as non-toxic (NT). When considering an example post \textit{I hate men}, we know that this post is both toxic and directed at men only; however, had this example post been in the training set, it would have erroneously taught the women branch of the baseline MTL models that the post was non-toxic. Similarly, an example post \textit{I hate women}, which is toxic and targeted at women only, would have contaminated the weights of the men branch of the baseline MTL models by skewing it to make more non-toxic predictions. While this weight-skewing effect of label contamination may result in higher accuracy scores for TradMTL and CSMTL because the majority of the Wilds dataset is nontoxic (85\% of the dataset is non-toxic), these models will subsequently acquire a poor understanding of toxicity. Conversely, CondMTL ensures that the demographic group branches learn a more accurate understanding of toxicity and correctly labels more toxic posts as toxic, as illustrated by higher group-specific recall values on the toxic posts in the testing dataset. 

% Numerically, we can observe CondMTL's improvements in \textbf{Table \ref{tab:wilds-results}}; in particular, we highlight the significant improvement in recall for the group-specific branches for toxic (T) posts. In the context of automatic TL detection, we would like to ensure that toxic posts are not mislabeled as nontoxic, as such errors would disproportionately affect marginalized demographic groups. A higher recall value of toxic posts implies a model's improved capability to detect toxic posts. As a result, evaluating models using F1, precision, and recall is preferable to comparing their accuracies, as they better capture the models' capability to detect TL.

% The discrepancy between the F1, Recall, and Precision scores for the CondMTL model when compared to the Traditional MTL scores for the men and women demographic-specific branches highlights the benefits of our proposed labeling schema and conditional loss function. While the traditional MTL model suffers from a misleading labeling schema, which causes the model to calculate and backpropagate misleading errors, CondMTL correctly identifies examples relevant to each demographic branch, resulting in drastically improved numbers ($\sim38$ for F1, $\sim29$ for Recall and $\sim56$ for Precision) compared to ($\sim8$ for F1, $\sim4$ for Recall and $\sim48$ for Precision) for the men branch and ($\sim40$ for F1, $\sim31$ for Recall and $\sim54$ for Precision) compared to ($\sim5$ for F1, $\sim3$ for Recall and $\sim47$ for Precision) for the women branch.

\textbf{Measures of algorithmic fairness and their usage.} Frequently, models that seek to improve algorithmic fairness do so by directly considering a fairness measure as part of the loss function, which is often done in the form of a penalty term. In our optimization objective, we do not incorporate any algorithmic fairness measure. We use a variant of weighted Binary Cross Entropy (wBCE) for optimizing the MTL model branches which correlates to giving higher priority in detecting examples from the smaller toxic class. Rather than modifying the loss function as a result of our fairness concerns, we modify the network architecture and labeling schema in a way that enables us to better capture heterogeneity across groups. This approach is suitable for settings in which improving recall for the minority group is a primary fairness consideration. However, if the primary concern is the \emph{difference} across groups, the proposed approach may not always yield improvements, because even if recall improves for both groups, the improvement could be greater for the majority class. In such a scenario, we would like to optimize the network \wrt a fairness measure, and thus we need to use a differentiable version of that said fairness measure. %While we do not use any optimization measure that correlates to minimizing Equal Opportunity, 
There exist works in the literature \citep{shen2022optimising} that can optimize a network for a fairness measure. Correspondingly, networks can also be optimized for equal accuracy across groups \citep{kovatchev2022fairly} or equalized odds \citep{roh2021fairbatch}. The choice of the measure depends on the practitioner's need and the availability of a differentiable version of said measure.

When considering intersectional fairness, \eg the intersectionality of gender and race, or a more fine-grained grouping of demographics, the dimensions of groups increase. In these cases, the performance of decoupled approaches drops due to data sparsity. We anticipate that the benefits of the shared layer in CondMTL would be even more salient in this setting.
% \textbf{Differential subgroup validity and heterogeneity across groups} In the absence of differential subgroup validity, learning one model for all groups would be the best solution, 
% \textbf{We need a paragraph here to talks about DSV \wrt our results} Need a skeleton to write around it. \textbf{HELP!!!}

\textbf{Other stakeholders in toxicity detection.} When considering toxicity detection, there are multiple stakeholders who are involved. We have primarily focused on the subject of the post, but other stakeholders include the author of the post and the annotator. Previous work has shown risks of algorithmic bias affecting authors of posts; for instance ~\citep{sap2019risk} shows that models may exhibit disporportionately high false positive rates for posts written in African American English. The importance of considering the demographics of annotators involved in labeling data has also been recently emphasized~\citep{sachdeva2022assessing}. Using the proposed CondMTL to model the problem in relation to other stakeholders' demographics is a natural extension of the proposed work.% A conceptual extension of our proposed work is framing fairness in toxicity detection as encompassing a set of related but distinct problems, rather than any single, monolithic problem. Problem 1: {\em Target Perspective} -- models should ensure fair protection across different demographic groups targeted by toxicity. Problem 2: {\em Author Perspective} -- models should provide fair moderation of posts written by authors from different demographic groups. Problem 3: {\em Community Perspective} -- different demographic groups often have different standards for what constitutes toxicity. Rather than learning a single, one-size-fits-all detection model for all groups, models should flexibly fit varying community standards for toxicity. In this work, we showed results for Problem 1 (\textit{target fairness}) only.

\textbf{Extension to other problem domains.} 
%\hl{When considering the intersectionality of gender and race or a more fine-grained grouping of demographics, the dimensions of groups increase. In these cases, the performance of decoupled approaches drops due to data sparsity. We anticipate that the benefits of the shared layer in CondMTL would be even more salient in this setting.} Furthermore, 
The proposed CondMTL can be easily applied to tackle group specific labeling in domains such as media bias detection and fact-checking, where sparsity in dataset labels could similarly lead to label contamination. 

% \textbf{Paragraph around some test posts and their predictions on each of the models, following template schema in Table 2}

% highly contextual -> the model has seen many training examples including the GOP but likely
%  does not understand what the GOP is / or that it is directed at both men and women

\vspace{-0.5em}
\section{Conclusion}

In this work, we frame the challenge of demographic-specific toxicity detection as a multi-task learning problem. We also note that the traditional MTL labeling schema causes label contamination under this problem setting, since posts targeted towards one group are labeled in a way that is misleading to the loss of branches specialized in other groups. To bypass this issue and address differential subgroup validity in the context of toxic language, we propose both: a) an updated labeling schema to avoid label contamination; and b) a conditional MTL framework with group-specific loss function based on a selective binary cross entropy formulation. The proposed architecture is shown to have significantly fewer trainable parameters and runtime than the stacked single task baseline, making it more suitable for deployment. Finally, we experimentally demonstrate that our framework achieves higher group recall values for the smaller toxic class over other baselines, and for the minority demographic group in particular, at no cost in overall accuracy compared to other MTL models.

%\vspace{-0.5em}
\section*{Acknowledgments}
We thank the reviewers for their valuable feedback. This research was supported in part by Wipro, an Amazon Research Award, and by {\em Good Systems\,}\footnote{\scriptsize\url{http://goodsystems.utexas.edu/}}, a UT Austin Grand Challenge to develop responsible AI technologies. Our opinions reflect our own views only. % and not the views of the sponsoring agencies.
%}

\clearpage

% \balance
\bibliography{References}

\clearpage

\appendix

\section{Experimental Details} \label{app:setup}

\subsection{Setup}

Experiments use a Nvidia 2060 RTX Super 8GB GPU, Intel Core i7-9700F 3.0GHz 8-core CPU and 16GB DDR4 memory. We use the Keras \citep{chollet2015} library on a Tensorflow 2.8 backend with Python 3.7 to train the networks in this paper. For optimization, we use AdaMax \citep{kingma2014adam} with parameters (\textit{lr}=1e-5) and $1000$ steps per epoch. For each configuration, we did five independent runs to report mean and variance across different baselines and our method.

\subsection{Training Loss Trajectory}

\begin{figure}[ht]
    \centering
    \vspace{-0.5em}
    \includegraphics[width=0.6\linewidth]{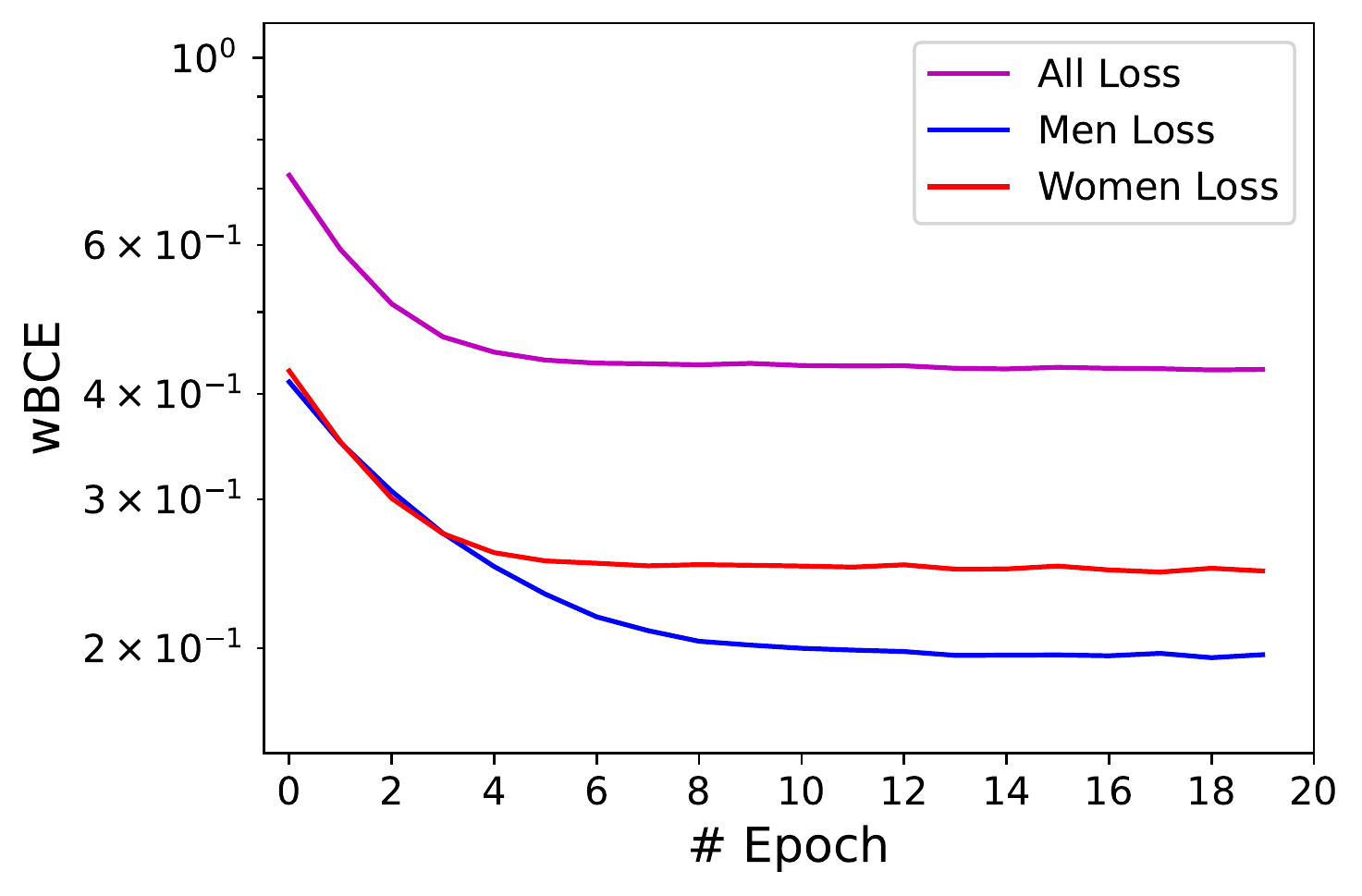}
    \vspace{-0.5em}
    \caption{\small Loss trajectory over training data for CondMTL for 20 epochs. Note that due to the conditional nature of the formulation, the error in the men (minority group) goes down as well.}
    \vspace{-1em}
    \label{fig:loss_traj}
\end{figure}

We show the loss trajectory over the training data in \textbf{Fig. \ref{fig:loss_traj}}. One can observe that the use of AdaMax results in the classification loss \ie weighted Binary Cross Entropy (WBCE) dropping rapidly at first and then stabilizing at around 10 epochs. Since the All branch deals with the full dataset, it achieves the highest error value. Due to the conditional nature of CondMTL, which only considers group relevant examples for each branch, we see improved loss over both the groups (men and women). More specifically, both groups start roughly at the same loss, with the men group achieving lower error values after stabilizing. This empirically highlights that CondMTL is not biased towards the majority group (women), rather giving importance to the minority group (men) as well.

\subsection{Class Balancing Strategy}

As common in most toxic language tasks, the Wilds dataset has a class label imbalance, with most ($85\%$) of the posts belonging to the non-toxic class. To account for this, we perform a class balancing strategy to assign more weight to the toxic examples during model training. We use a weighted version of Binary Cross Entropy (BCE) measure that re-weights the error for the different classes proportional to their inverse frequency in the data \citep{lin2017focal}. 

We use SKLearn's \citep{scikit-learn} \textit{compute\_class\_weights=`balanced'} flag for extracting weights ($w_{\textrm{toxic}}, w_{\textrm{non-toxic}}$) for toxic and non-toxic classes for each branch, given as\footnote{\url{https://scikit-learn.org/stable/modules/generated/sklearn.utils.class_weight.compute_class_weight.html}}: 
\begin{equation}
    n\_samples / (n\_classes \times bincount(y))
\end{equation}

\section{Mean Variance Numbers} \label{app:variance}

We report mean and standard deviation results of five runs over the model by using wBCE loss function. For simplicity, we show the overall accuracy numbers in Table \ref{tab:wilds-acc}. We observe that all of the models roughly perform the same \wrt the Stacked STL ($\sim 86\%$) and with negligible variance over runs. Note that all the five runs were independent \ie their network weights were randomly initialized at start of each training run.

\begin{table}[ht]
% \vspace{-1.5em}
\centering
    \resizebox{0.8\linewidth}{!}{%
	\begin{tabular}{l|r|r|r} \toprule
	Loss type & All & Men & Women \\ \midrule
        Stacked Single Task & 86.3 $\pm$ 0.1 & 85.6 $\pm$ 0.2 & 87.0 $\pm$ 0.1 \\ \midrule
			Cross Stitch Multi Task \citep{misra2016cross} & 86.3 $\pm$ 0.0 & 85.8 $\pm$ 0.1 & 86.6 $\pm$ 0.0 \\
			Traditional Multi Task & 86.2 $\pm$ 0.2 & 85.5 $\pm$ 0.2 & 86.7 $\pm$ 0.1 \\ \midrule
			Conditional Multi Task (Ours) & 86.2 $\pm$ 0.2 & 85.7 $\pm$ 0.1 & 86.7 $\pm$ 0.1 \\ \bottomrule
	\end{tabular}
	}
% 	\vspace{-0.5em}
	\caption{\small Mean and Standard Deviation measures of models across five runs. All the models perform the same \wrt Overall Accuracy.}
	\label{tab:wilds-acc}
    \vspace{-1.5em}
\end{table}

\section{Label Contamination} \label{app:contam}

\begin{figure}[ht]
    \centering
    \begin{subfigure}{0.4\linewidth}
        \includegraphics[width=\linewidth,page=1]{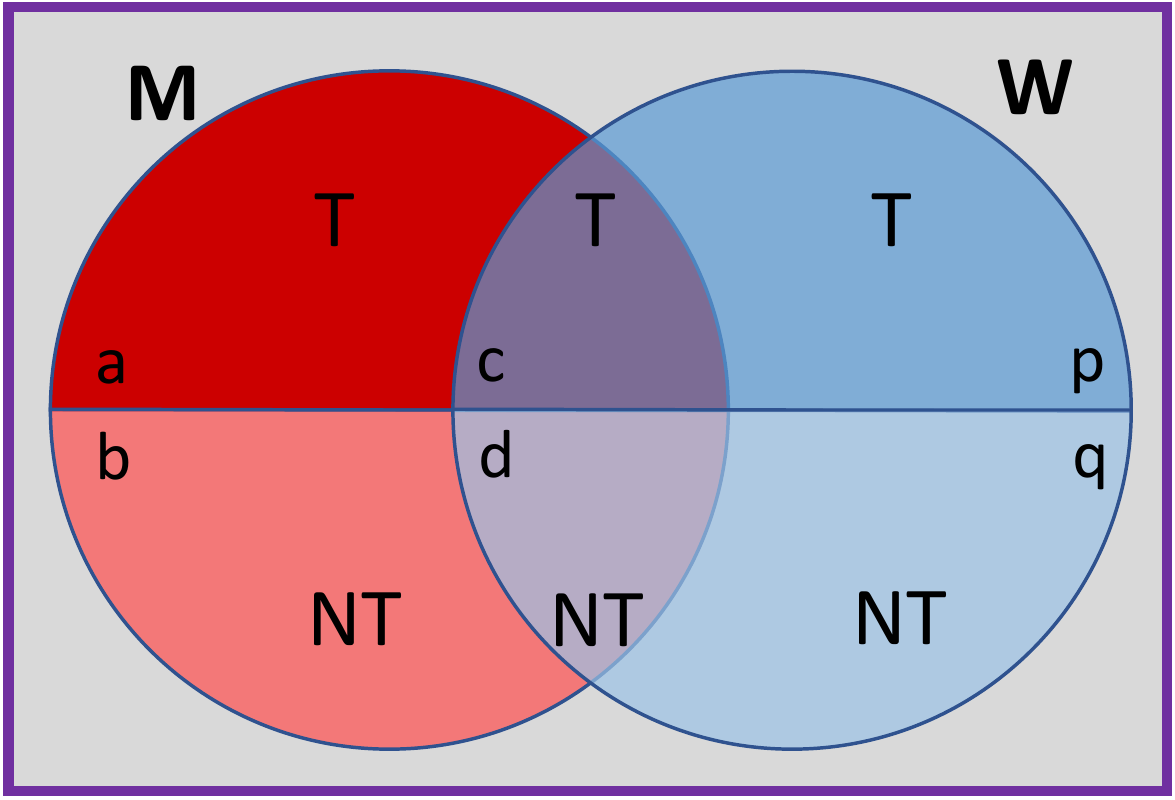}
        \vspace{-1.5em}
        \caption{\small True Labels}
    \end{subfigure}
    \quad
    \begin{subfigure}{0.4\linewidth}
        \includegraphics[width=\linewidth,page=2]{figs/www23.pdf}
        \vspace{-1.5em}
        \caption{\small TradMTL Label for Men}
    \end{subfigure}
    \vspace{-1em}
    \caption{\small Illustration of label contamination when following TradMTL labeling schema due to false consideration of group-irrelevant posts as non-toxic (NT). ($a$) shows the true labels for each group, being toxic/non-toxic. However, following the TradMTL scheme for group (M) results in both toxic/non-toxic posts explicitly from group (W) (regions $p$ and $q$) to falsely be grouped with the non-toxic labels for group (M) (ideally just regions $b$ and $d$).}
    \vspace{-0.5em}
    \label{fig:contam}
\end{figure}

To illustrate \textit{label contamination}, let us consider a toxicity labeling task with binary labels (NT and T) and binary groups (M and F). Given the set of all possible posts, indicated by the grey rectangle in \textbf{Fig. \ref{fig:contam} (a)}, the set of posts belonging to the groups M and F are indicated by their respective circles, red ($\mathcal{D}_1$) and blue ($\mathcal{D}_2$) shades respectively. The combined dataset $\mathcal{D}$ is a union of the two circles ($\mathcal{D}=\mathcal{D}_1 \cup \mathcal{D}_2$) with some post targeted towards individual groups, and some targeting both groups (intersection). Within each group, we have a set of toxic (T) \vs non-toxic posts, indicated by dark and light shades respectively. The individual portions of the venn diagram are marked with lower case alphabet for clarity, as \eg region ($a$) indicates toxic post targeted towards men only, while region ($c$) indicates toxic posts targeted at both men and women. 

\textbf{Table \ref{tab:trad_schema}} shows the ideal and cross contaminated labels for posts \wrt toxic \vs no-toxic labels pertaining to each group. Due to label contamination in TradMTL, the non-toxic parts $NT|M$ and $NT|F$ have additional group-irrelevant labels associated with them which causes the classifier to learn incorrect boundaries.

\begin{table}[ht]
    \centering
    \begin{tabular}{l|r|r|r|r} \toprule
        Group & T | M & T | F & NT | M & NT | F \\ \midrule
        True Label & $a \cup c$ & $c \cup p$ & $b \cup d$ & $d \cup q$ \\
        TradMTL Label & $a \cup c$ & $c \cup p$ & $b \cup d \, {\color{red} \cup \, p \cup q}$ & $d \cup q \, {\color{red} \cup \, a \cup b}$ \\ \bottomrule
    \end{tabular}
    \caption{\small TradMTL schema assumes group-irrelevant posts, leading to mislabeling in each non-toxic group, marked in red.}
    \vspace{-2.0em}
    \label{tab:trad_schema}
\end{table}

\section{Benchmarks} \label{app:benchmark}

We verify our stated propositions for CondMTL and CSMTL through simple and verifiable benchmarking cases on a regression and classification task. We consider the integer group $\mathbb{Z}$ of numbers for the tasks \{1,2,\ldots,10\}, with two groups (A and B), where group [A] contains the numbers \{1,2,\ldots,5\} and group [B] contains the numbers \{6,7,\ldots,10\}. To generate the actual data and mimic variability and imbalance, we replicate each number in the set in increments of 100 \ie $\mathcal{D}=[1:100, 2:200, \ldots, 10:1000]$ with each group data as $\mathcal{D}_A=[1:100, 2:200, \ldots, 5:500]$ and $\mathcal{D}_B=[6:600, 7:700, \ldots, 10:1000]$. The data notation $d=[p:q]$ indicates that there are $q$ instances of element $p$ in the dataset.

\subsection{Regression Task}

For the regression task, we check each MTL model's performance on a multiplication task. We define the following three tasks: T1) given data $X$ produce $Y=4X$ over the dataset $\mathcal{D}$; T2) given data $X$ produce $Y=2X$ over the dataset $\mathcal{D}_A$; and T3) given data $X$ produce $Y=6X$ over the dataset $\mathcal{D}_B$. The architecture contains a shared layer of 4 dense neurons and two depths for the task specific neurons containing 2 and 1 neurons respectively. Since scaling a number is a linear operation, we keep the dense neurons in the network with linear activations, and mean squared error loss. 

\begin{figure}[ht]
    \centering
    \begin{subfigure}{\linewidth}
        \includegraphics[width=\linewidth]{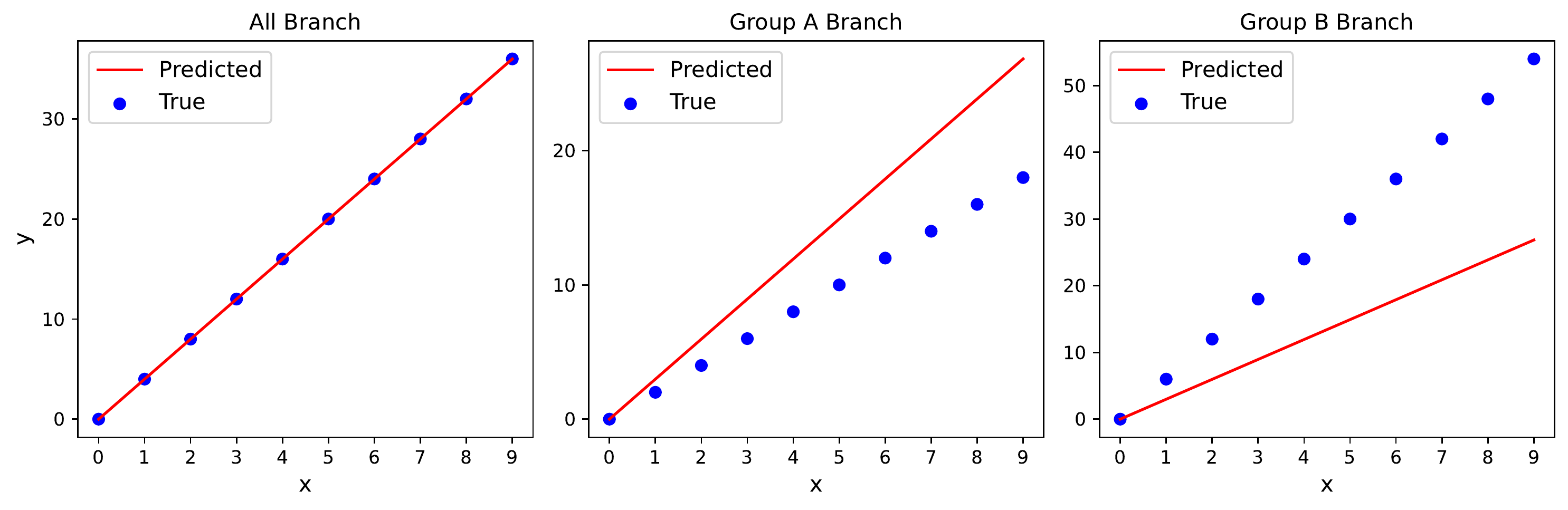}
        \vspace{-2.0em}
        \caption{\small Predictions for CSMTL}
    \end{subfigure}
    % \quad
    \begin{subfigure}{\linewidth}
        \vspace{1em}
        \includegraphics[width=\linewidth]{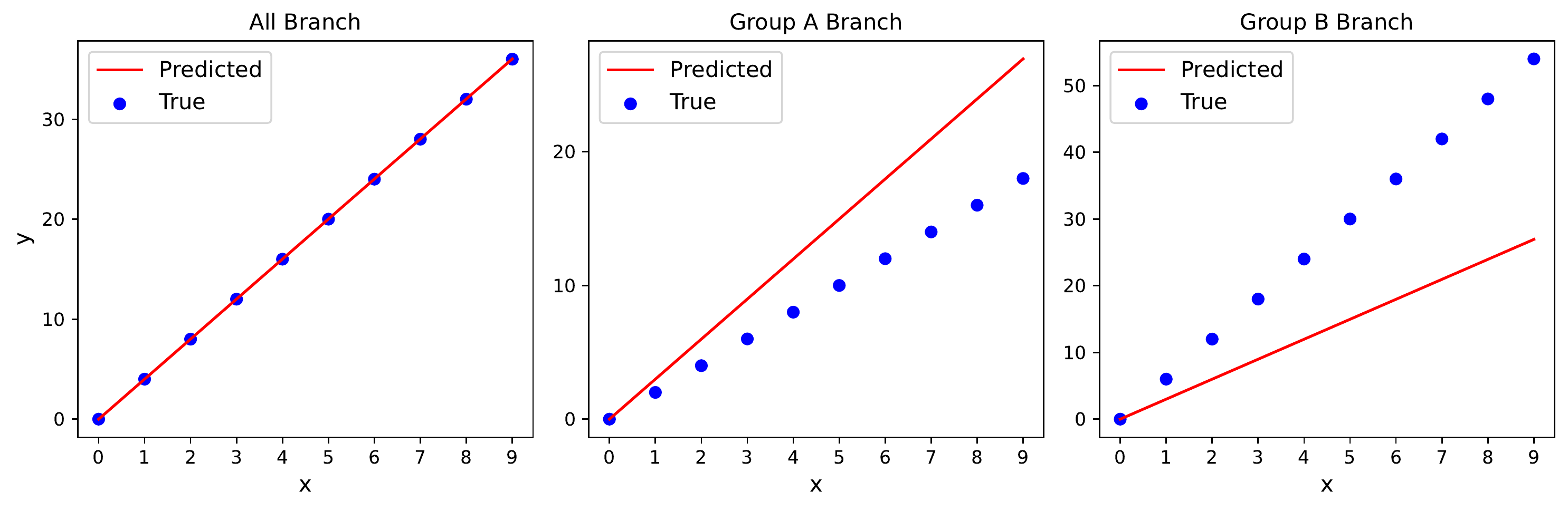}
        \vspace{-2.0em}
        \caption{\small Predictions for TradMTL}
    \end{subfigure}
    \begin{subfigure}{\linewidth}
        \vspace{1.0em}
        \includegraphics[width=\linewidth]{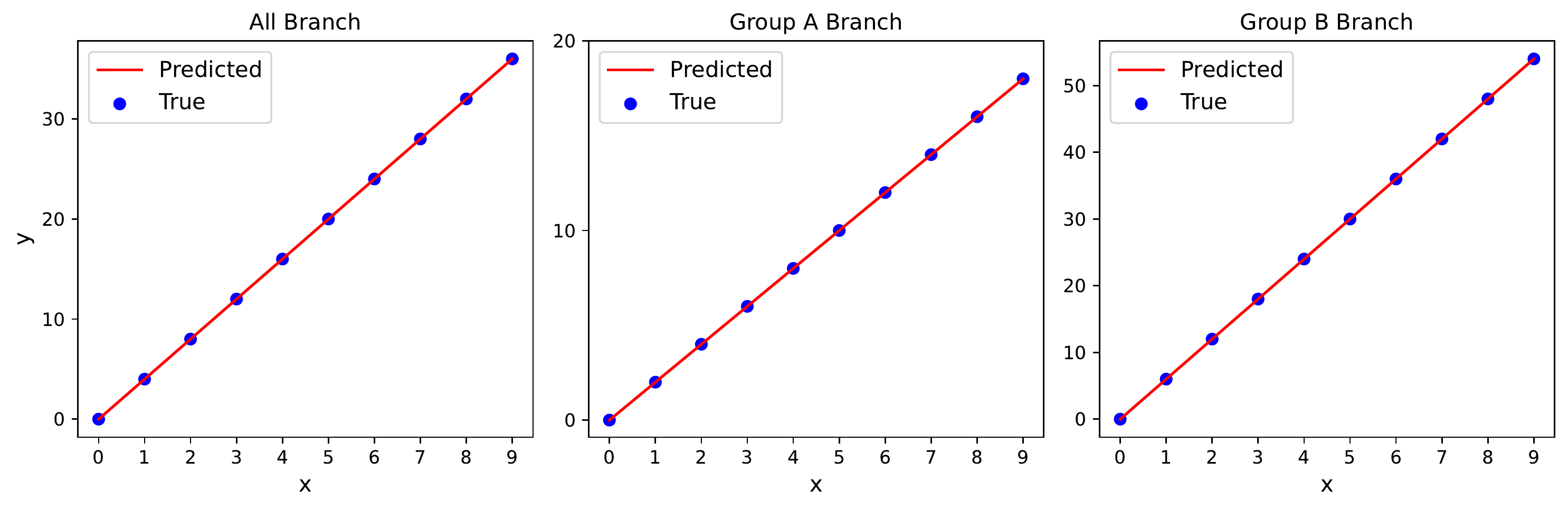}
        \vspace{-2.0em}
        \caption{\small Predictions for CondMTL}
    \end{subfigure}
    \vspace{-1.5em}
    \caption{\small Predictions of the three MTL models over the benchmark regression tasks, where T1) $y=4x$, T2) $y=2x$ and T3) $y=6x$. Since the all branch looks at the full data $\mathcal{D}$, it performs accurately over all the models. However, for the Group [A] and [B] branches, the CSMTL and TradMTL learns from a mix of labels of $\mathcal{D}_A$ and $\mathcal{D}_B$, which leads to its predictions overshooting and undershooting respectively.}
    \vspace{-1.0em}
    \label{fig:bench_regr}
\end{figure}

Fig. \ref{fig:bench_regr} shows the results obtained by the MTL models, where due to label contamination, both the CSMTL and TradMTL suffer in predictions for Groups A and B. Since CondMTL makes branch [A] learn over examples of $\mathcal{D}_A$ explicitly, it learns the representation of $Y=2X$ and is able to predict $2X$ both in and out of domain of $\mathcal{D}_A$. Same logic holds for branch [B] which can predict $6X$ both in and out of domain of $\mathcal{D}_B$. Also looking at the entries of the CS matrix, one can observe that both identity and symmetry does not hold.
\begin{align*}
    CS = \begin{bmatrix*}[r]
    1.20 & 0.41 & -0.34 \\
    0.47 & 1.21 & -0.11 \\
    -0.43 & -0.46 & 1.29
    \end{bmatrix*}
\end{align*}

In another regression variant, we intentionally set task T2 as: given data $X$ produce $Y=0$ over the dataset $\mathcal{D}_A$, which essentially implies predicting a constant value for all input data. This updated task T2 is independent of both tasks T1 and T3, where the Cross Stitch unit should not communicate anything through T2. However, after training, we still observe that the off-diagonal CS matrix entries corresponding to Task 2 are still non-zero leading to contamination of weights across other task layers.

\subsection{Classification Task}

For the classification task, we keep the same architecture, but replace linear activation with \textit{tanh} and for the final layer a \textit{sigmoid} activation. We define the following three tasks: T1) given data $X$ classify $Y=1$ if $4 \leq X \leq 7$ else $0$ over the dataset $\mathcal{D}$; T2) given data $X$ classify $Y=1$ if $X \geq 4$ else $0$ over the dataset $\mathcal{D}_A$; and T3) given data $X$ classify $Y=1$ if $X \geq 7$ else 0 over the dataset $\mathcal{D}_B$. Similar faulty performances occur in this setting as well in both CSMTL and TradMTL due to label contamination, while CondMTL learns the correct group specific representation.

\end{document}